\documentclass{article}

\usepackage[preprint]{neurips_2026}

\usepackage[utf8]{inputenc} 
\usepackage[T1]{fontenc}    
\usepackage{hyperref}       
\usepackage{url}            
\usepackage{booktabs}       
\usepackage{amsfonts}       
\usepackage{nicefrac}       
\usepackage{microtype}      
\usepackage{xcolor}         

\usepackage{booktabs}     
\usepackage{multirow}     
\usepackage{graphicx}     
\usepackage{array}  
\usepackage{amsmath}
\usepackage{subcaption}
\usepackage{enumitem}

\title{UniBCI: Towards a Unified Pretrained Model for Invasive Brain-Computer Interfaces}

\author{%
  Binjie Hong\footnotemark[1] \\
  Brain Science Data Center\\
  CEBSIT, Institute of Neuroscience\\ Chinese Academy of Sciences\\
  \texttt{hongbj@ion.ac.cn} \\
  \And
  Rui Xiong\footnotemark[1] \\
  Brain Science Data Center\\
  CEBSIT, Institute of Neuroscience\\ Chinese Academy of Sciences\\
  \texttt{xiongrui@ion.ac.cn} \\
  \AND
  Liyuan Han \\
  CEBSIT, Institute of Neuroscience\\ Chinese Academy of Sciences\\
  \texttt{hanliyuan@ion.ac.cn} \\
  \And
  Tielin Zhang\footnotemark[2] \\
  CEBSIT, Institute of Neuroscience\\ Chinese Academy of Sciences\\
  \texttt{zhangtielin@ion.ac.cn} \\
}

\begin{document}

\footnotetext[1]{Equal contributions}
\footnotetext[2]{Corresponding author}

\maketitle

\begin{abstract}
Modeling invasive neural spike data is fundamental to advancing high-performance brain-computer interfaces (BCIs). However, existing approaches face critical challenges, including limited-scale heterogeneous data, cross-domain distribution shift, and the intrinsic spatiotemporal complexity of invasive neural signals. In this work, we propose \textbf{UniBCI}, a \textbf{uni}fied pretrained model for \textbf{i}nvasive \textbf{B}rain-\textbf{C}omputer \textbf{I}nterfaces. The model integrates three key components: (1) a \emph{context-conditioned spatio-temporal tokenization (CST)} scheme that embeds neural signals together with metadata into a shared representation space; (2) a hierarchical \emph{Interval-Area Attention (IAA)} mechanism that captures patterns of spike dynamics in slots via linear attention and locality dependencies via sliding-window attention; and (3) a scalable \emph{self-supervised masked signals reconstruction} objective for learning generalizable neural representations from large-scale unlabeled data. We construct a pretraining corpus spanning multiple species, subjects, brain regions, and behavioral experiment paradigms. These heterogeneous recordings are standardize via our proposed unified normalization and tokenization. Comprehensive experiments demonstrate that UniBCI achieves SOTA performance across diverse downstream tasks while improving generalization. Moreover, the model achieves a strong balance between accuracy and efficiency, with fewer trainable parameters and lower inference latency. These results suggest that UniBCI provides a practical step toward general-purpose neural foundation models, enabling robust, scalable, and transferable representation learning for invasive neural data. The code for this paper is available at: \url{https://anonymous.4open.science/r/UniBCI-C805}.
\end{abstract}

\section{Introduction}
Brain information processing can be broadly categorized, based on the observation modality, into two levels: non-invasive and invasive neural signals. Non-invasive techniques, such as Electroencephalography (EEG) and functional Magnetic Resonance Imaging (fMRI) provide high safety levels and broad spatial coverage. They have achieved substantial progress in brain state decoding and cognitive inference, forming increasingly mature modeling paradigms. With the advent of large-scale datasets, they have benefited from the foundation model paradigm, improving cross-subject generalization, self-supervised pretraining, and multi-task learning ~\cite{wang2024eegpt, zhou2025csbrain, wang2025cbramod}. However, the limited spatial and temporal resolution of non-invasive signals constrains their ability to capture fine-grained spiking dynamics and microscopic neural processes. In contrast, invasive recording techniques (e.g., multi-electrode arrays~\cite{tsai2017very} and Neuropixels~\cite{jun2017fully}) enable high-resolution measurements at the level of individual neurons and spikes. They provide a critical substrate for studying neural computation and developing invasive brain–computer interfaces (BCIs). 

Early work on invasive neural data modeling primarily relied on classical statistical and dynamical system methods, such as Wiener filters~\cite{wiener}, Kalman filters~\cite{Kalman}, and Gaussian-Process Factor Analysis (GPFA)~\cite{GPFA}. Although these approaches can capture low-dimensional latent structures, they depend on strong assumptions such as linearity and stationarity, limiting their ability to model complex nonlinear neural dynamics. With the rise of deep learning, research has shifted toward approaches based on recurrent neural networks (RNNs)~\cite{RNN}, Transformers~\cite{Transformers}, and state space models (SSMs)~\cite{SSM}, which better capture temporal dependencies and improve performance in motor decoding and behavioral prediction. However, these models often suffer from high data requirements, limited interpretability, and poor generalization across sessions or subjects. More recently, large-scale pretraining paradigms have emerged, which can be broadly divided into self-supervised and behavior-supervised approaches. Representative methods such as NDT1~\cite{ye2021ndt} introduce Transformer-based spike modeling, while NDT2~\cite{NDT2} extends this framework to cross-session, cross-subject, and multi-context pretraining via spatiotemporal transformers. Although NDT2 alleviates the limitations of NDT1, it introduces substantial computational overhead and scaling complexity. Behavior-supervised researches such as NDT3~\cite{ndt3} further scale this paradigm to hundreds of datasets. In addition, POYO~\cite{poyo} and POSSM~\cite{POSSM} introduce learnable neural token representations to enable scalable multi-subject pretraining, but rely heavily on large amounts of high-quality behavioral annotations, which are costly and often unavailable. Despite these advances, existing methods remain limited to single-species settings and lack unified modeling across brain regions, species, and tasks, leaving general-purpose neural modeling an open problem.

This limitation is further amplified in invasive brain–computer interface (BCI) and closed-loop neuroscience applications, which require simultaneous high accuracy, low latency, and strong generalization~\cite{vermani2024real}. For instance, in motor decoding~\cite{flesher2021brain}, speech prostheses~\cite{willett2023high}, and closed-loop neuromodulation~\cite{bonizzato2023autonomous, lorach2023walking}, systems must operate reliably in real-time interactive environments. However, current pretraining models typically exhibit trade-offs among these objectives, and no unified neural foundation model achieves all three simultaneously. This gap is concluded as three key challenges: data heterogeneity with limited scale, cross-domain distribution shift, and spatiotemporal complexity of invasive signals.

We propose a unified neural foundation modeling framework inspired by intrinsic characteristics of invasive brain signals. \textbf{First}, to address dataset fragmentation and massive data without behavioral labels, we construct a unified data framework with a multi-source pretraining corpus.  \textbf{Second}, we introduce a context-conditioned spatio-temporal tokenization (CST) strategy. The metadata consisting of species, subjects, brain regions, tasks, and experimental paradigms is encoded as context embeddings and injected into spike representations, enabling conditional modeling within a unified parameter space and improving robustness to domain shift. \textbf{Third}, Interval-Area Attention (IAA) is innovated to model neural dynamics at different levels of granularity. Within each interval, linear attention captures spiking patterns within a slot, and area-wise sliding-window attention models population dynamics among adjacent brain areas. This architecture enables joint modeling of temporal and spatial neural representations while improving computational efficiency. We adopt a self-supervised masked spike reconstruction objective to learn general neural representations from unlabeled heterogeneous data. Our comprehensive experiments demonstrate SOTA performance on downstream datasets spanning multiple conditions. 

Contributions of this paper:
\begin{itemize}
    \item Unify massive invasive BCI data by normalization of raw spike trains and tokenization combined with metadata, enforcing spike representations to be fully expanded in a hidden space. 
    \item Propose UniBCI, a novel framework inspired by the characteristics of invasive signals, considering both temporal spikes and spatial dependency in area-range.
    \item Develop self-supervised methods of reconstructing spike signals, enhancing the generalization ability of representing neural data across multiple species, subjects, brain regions and task paradigms.
    \item Conduct comprehensive downstream experiments, demonstrating SOTA performance under both classification and regression tasks while showing high inference efficiency with fewer trainable parameters.
\end{itemize}

\section{Approach}

We propose UniBCI, a unified spatio-temporal foundation model for neural spike data that integrates structured preprocessing, context-aware tokenization, and hierarchical attention layers. Spike trains from different datasets have varying shapes, where the dimension of timestep and channel is diverse. We suppress temporal information by binning timesteps and group channels to normalize multi-source spike trains (Section \ref{2.1}). Normalized spike trains combined with metadata contexts are chunked into tokens, and then the interval-area attention is employed in a stacked architecture. During pretraining, we utilize the self-supervised method of reconstructing masked tokens (Section \ref{2.2}).

\subsection{Spike Train Normalization}
\label{2.1}
Raw neural spike recordings are irregular in both temporal sampling and channel organization. We transform them into a structured tensor representation. Let ${X}_\mathrm{raw} \in \mathbb{R}^{T_\mathrm{raw} \times C_\mathrm{raw} }$ denote the raw spike matrix, where $T_\mathrm{raw}$ is the original temporal resolution and $C_\mathrm{raw}$ is the number of channels. Each element $X_{\tau,c}$ represents the spike activity of the channel $c$ at time index $\tau$. During the temporal binning, we discretize time into bins and aggregate spikes:
\begin{equation}
\tilde{X}_{t,c} = \sum_{\tau \in \mathcal{B}(t)} X_{\tau,c},
\end{equation}
where $\mathcal{B}(t)$ is the set of timestamps in bin $t$. This maintains temporal information of spike activities, producing $\tilde{X} \in \mathbb{R}^{{T_\mathrm{norm} \times C_\mathrm{raw} }}$. Considering the spatial relationship in adjacent neurons, channels are grouped into $A$ areas $\{{C}_a\}_{a=1}^A$. The signals in a local area are denoted as
\begin{equation}
    \hat{{X}}_{t,a} =
\left[
\tilde{{X}}_{t,c_1}
\;\|\;
\tilde{{X}}_{t,c_2}
\;\|\;
\cdots
\;\|\;
\tilde{{X}}_{t,c_{|\mathcal{C}_a|}}
\right],
\end{equation}
where $|{C}_a|$ denotes the number of channels in the area $a$. We then truncate or pad channels in each area to a fixed number $C_\mathrm{norm}$, resulting normalized spike trains ${X}_{\mathrm{norm}} \in \mathbb{R}^{T_{\mathrm{norm}} \times A \times {C}_\mathrm{norm} }$.

\begin{figure}[t]
    \centering
    \includegraphics[width=1\linewidth]{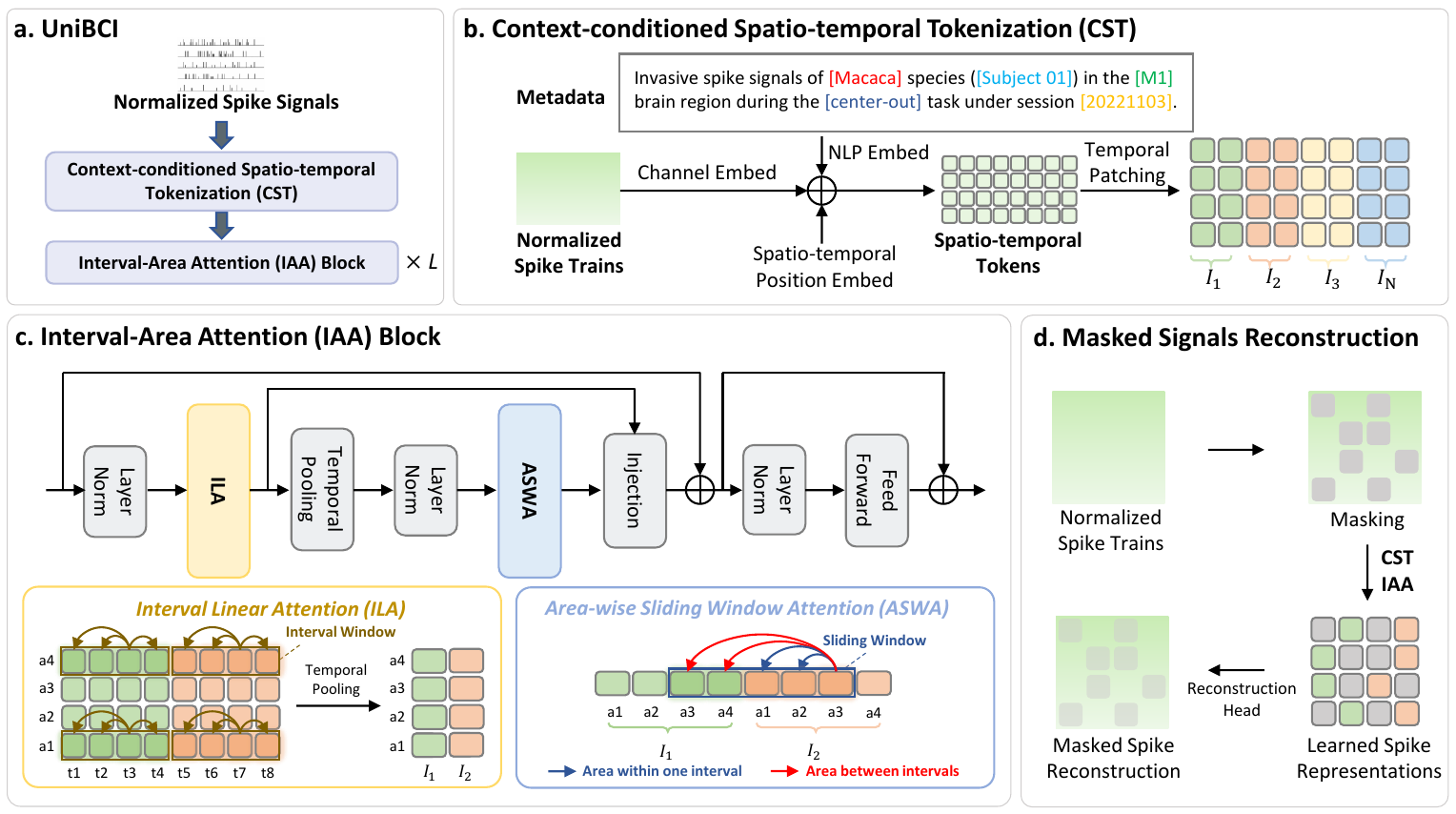}
    \caption{\textbf{a.} Architecture of UniBCI consists of context-aware tokenization and stacked interval-area attention layers. \textbf{b.} The normalized spike signal is tokenized with metadata. \textbf{c.} Then interval embeddings are through interval linear attention $ \mathrm{ILA}$ to learn temporal features within a short slot. Followed by area-wise sliding window attention $\mathrm{ASWA}$, spatial dependency of spikes is considered. \textbf{d.} The reconstruction objective aims to generate masked parts of spike tokens from extracted features of spike trains.}
    \label{architecture}
\end{figure}
\subsection{UniBCI}
\label{2.2}
Inspired by the characteristics of invasive neural activity, we propose UniBCI, which reconstructs masked spike tokens by optimizing the tokenization module and the interval-area attention blocks. This is illustrated in Figure \ref{architecture}. In the tokenization stage, normalized spike trains are chunked into spatio-temporal tokens by incorporating contextual representations. The interval and area-wise attention blocks that consider temporal and locality dependency are stacked to extract meaningful features of spike signals. 

\paragraph{Context-conditioned Spatio-temporal Tokenization (CST):} Given ${X}_{\mathrm{norm}} \in \mathbb{R}^{T_{\mathrm{norm}} \times A \times {C}_\mathrm{norm} }$, we propose a tokenization strategy to serialize spike trains by integrating contexts and spatio-temporal positions. Specifically, a set of learnable weights  ${W}_e\in \mathbb{R}^{ C_{\mathrm{norm}} \times d}$ and ${b}_e \in \mathbb{R}^{d}$ is shared among $A$ areas, encoding channels into $d$-dimensional vectors:
\begin{equation}
X_{\mathrm{emb}} = \big\|_{a=1}^{A}({X}_a\cdot {W}_e  + {b}_e),
\end{equation}
where $X_{\mathrm{a}} \in \mathbb{R}^{T_\mathrm{norm}\times C_{\mathrm{norm}}}$ denotes spike recordings of $C_{\mathrm{norm}}$ channels and $T_{\mathrm{norm}}$ timesteps in the area $a$, resulting in $X_{\mathrm{emb}} \in \mathbb{R}^{A \times T_{\mathrm{norm}} \times d}$ as embedded spike signals. Since spike trains stem from diverse paradigms, the metadata of species, subjects, brain regions, tasks and session identities are incorporated as the template:

\textit{Invasive spike signals of {\color{red}[Species]} species ({\color{blue}[Dataset] [Subject]}) in the {\color{green}[Region]} brain region during the {\color{purple}[Task]} task under session {\color{orange}[Session/Date]}.}

\begin{figure}[t]
    \centering
    \begin{minipage}{0.45\textwidth}
        \centering
        \includegraphics[width=\linewidth]{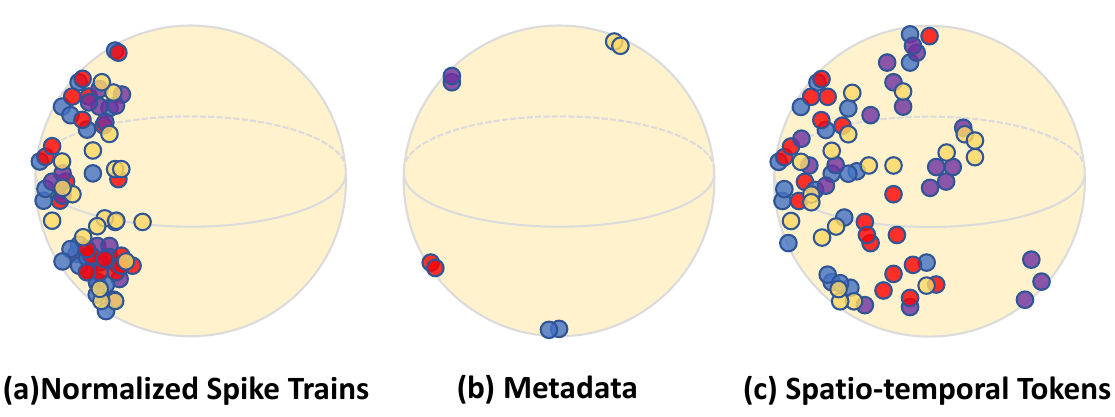}
        \caption{Context embeddings enrich representation of spike tokens, ensuring expanded embedding space of spatio-temporal tokens.}
        \label{text}
    \end{minipage}
    \hfill
    \begin{minipage}{0.45\textwidth}
        \centering
        \includegraphics[width=\linewidth]{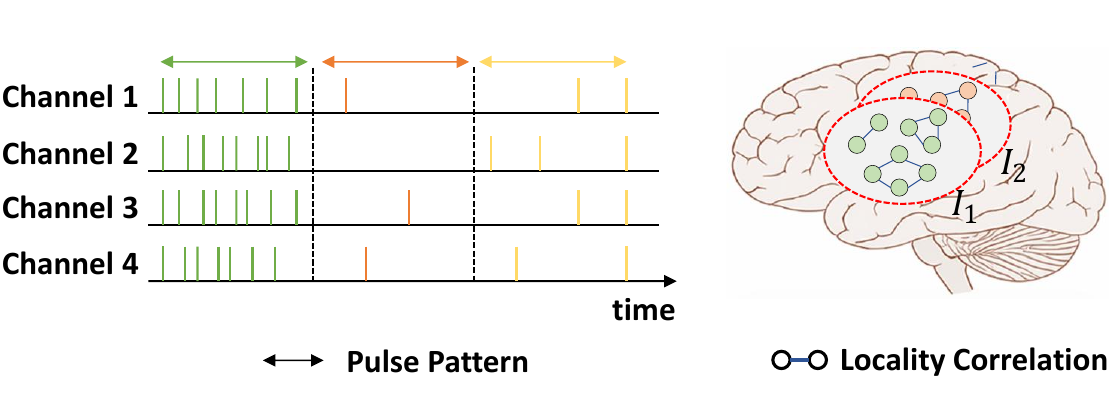}
        \caption{Spikes pulsing in a slot pattern differently and adjacent channels recorded in one area have locality correlation.}
        \label{spike}
    \end{minipage}
\end{figure}

Leveraging a pretrained natural language model $\mathrm{NLM}$, we embed contexts ${m}$ in a dense space as vectors $X_{\mathrm{meta}}  = \mathrm{NLM}({m}) \in \mathbb{R}^{d}$. Furthermore, we employ two sets of learnable weights for embedding temporal and spatial positional embeddings, defined as $T_\mathrm{pos} \in \mathbb{R}^{1 \times T_{\mathrm{norm}} \times d}$ and $A_\mathrm{pos} \in \mathbb{R}^{A\times 1\times d}$ respectively. With broadcast addition of the context embedding, the final ${token} \in \mathbb{R}^{T_\mathrm{norm} \times A\times d }$ is formulated as:
\begin{equation}
{token} = X_{\mathrm{emb}} \oplus X_{\mathrm{meta}} \oplus T_\mathrm{pos} \oplus A_\mathrm{pos} .
\end{equation}
By combining the embedding of metadata, the hidden space of spikes is expanded, illustrated in Figure \ref{text}. The semantic representations of metadata make the spatio-temporal tokens distinguishable from different sources. Eventually, we partition spike train tokens into $N$ non-overlapping intervals over the temporal dimension, and the interval size is denoted as $t$ :
\begin{equation}
{I}_i \in \mathbb{R}^{t \times A \times d}, \quad i=1,\dots,N.
\end{equation}

\paragraph{Interval-Area Attention (IAA):}  Inspired by the characteristics of invasive signals, we design a stacked architecture where each layer considers interval and area dependency. Shown in Figure \ref{spike}, the temporal patterning between local intervals is non-stationary \cite{brown2004multiple}, and the spatial grouping across channels is based on field correlation of neural populations \cite{buzsaki2004large,stevenson2011advances}. Hence, we denote ${H}^{\ell} \in \mathbb{R}^{N\times A \times t \times d}$ as spike signals recorded from all areas $A$ during all intervals $N$ at layer $\ell$. Since one interval of spikes is much shorter than whole span of brain activity, we choose the mechanism of linear attention \cite{katharopoulos2020transformers} to accelerate inference speed although it doesn't reduce computational complexity in our case of $t < d$. We compute the representation of spikes $\tilde{O}_{i,a}^{\ell} \in \mathbb{R}^{t \times d}$ in one area from one interval ${H}_{i,a}^{\ell} \in \mathbb{R}^{t \times d}$:
\begin{equation}
\tilde{O}_{i,a}^{\ell} = {H}^{\ell}_{i,a} {W}_ q^{\ell} \cdot (({H}^{\ell}_{i,a} {W}_k^{\ell})^\top \cdot {H}^{\ell}_{i,a} {W}_v^{\ell}) ,
\end{equation}
where ${W}_q^{\ell}, {W}_k^{\ell}, {W}_v ^{\ell}\in \mathbb{R}^{d \times d}$ are shared for $N$ intervals and $A$ areas. Concatenating embeddings of spikes generated from $A$ areas during $N$ intervals, the output of interval linear attention is denoted as $\mathrm{ILA}^{\ell} \in \mathbb{R}^{N \times A\times t\times d}$:
\begin{equation}
\mathrm{ILA}^{\ell} =
\big\|_{i=1}^{N}
\left(
\big\|_{a=1}^{A}
\tilde{{O}}_{i,a}^{\ell}
\right).
\end{equation}
Concerning the locality correlation, the method of sliding window attention \cite{beltagy2020longformer, zaheer2020big, child2019generating} is employed. The queried area receives nearby signals and activities in the same area from historical intervals. Furthermore, this limits the attention scope in a sliding window, reducing quadratic computational complexity. We denote
$\tilde{H}^{\ell} \in \mathbb{R}^{S \times d}$ as the input of the area-wise attention module where $S = N \times A$ after temporal pooling in interval size $t$ and layer normalization: 
\begin{equation}
    \tilde{H}^{\ell} = \mathrm{LN}(\mathrm{AvgPool}(\mathrm{ILA}^{\ell})).
\end{equation}

We compute the area-wise sliding window attention written as ${O}^{\ell} \in \mathbb{R}^{S \times d}$:
\begin{equation}
{O}^{\ell} = \mathrm{softmax}(\tilde{H}^{\ell} {W}_Q^{\ell} \cdot (\tilde{H}^{\ell} {W}_K^{\ell})^\top \odot M) \cdot \tilde{H}^{\ell} {W}_ V^{\ell} .
\label{swa}
\end{equation}
where $M \in \mathbb{R}^{S \times S}$ is the sliding window mask where $M_{ij} = 1$ if $i-w + 1 \le j \le i$, otherwise $M_{ij} = -\infty$. Next, we reshape the output as $\mathrm{ASWA}^{\ell} =\mathrm{Reshape}(O^{\ell}) \in \mathbb{R}^{N \times A\times d}$ . The output of the next layer is formulated as follows:
\begin{equation}
    \mathrm{Z}^{\ell} =\mathrm{ILA}^{\ell} \oplus \mathrm{ASWA}^{\ell} + {H}^{\ell},
\end{equation}
\begin{equation}
{H}^{\ell+1} = \mathrm{FFN}(\mathrm{LN}( \mathrm{Z}^{\ell})) +  \mathrm{Z}^{\ell}.
\end{equation}
The features extracted from the stacked layers are utilized to reconstruct the masked spike tokens.
\label{2.3}
\paragraph{Masked Signals Reconstruction:} The initial token sequence with length $J=N \times A\times t$ is denoted as $\mathbf{X} \in \mathbb{R}^{J\times d}$ . We sample a binary mask $\mathcal{M} \in \{0,1\}^J$, where $\mathcal{M}_j = 1$ indicates that the $j$-th token is masked. Hence, the masked tokens are formulated as:
\begin{equation}
    \tilde{\mathrm{X}} = \{\tilde{\mathrm{x}}_j\}_{j=1}^J, \quad 
\tilde{\mathrm{x}}_j =
\begin{cases}
\mathrm{x}_j, & \mathcal{M}_j = 0, \\
0, & \mathcal{M}_j = 1.
\end{cases}
\end{equation}
Followed by stacked interval-area attention blocks, the latent representation of spike train tokens is encoded as $\mathrm{H} = \mathrm{IAA}(\mathbf{\tilde{\mathrm{X}}})$. A lightweight reconstruction head $g_\phi(\cdot)$ maps the latent features back to the spike token space $\hat{\mathrm{X}} = g_\phi(\mathrm{H}) \in \mathbb{R}^{J\times d}$. The training objective is defined as a masked reconstruction loss, computed only over masked tokens:
\[
\mathcal{L}_{\text{rec}} = \frac{\sum_{j=1}^J \mathcal{M}_j \cdot \ell\big(\hat{\mathrm{x}}_j, \mathrm{x}_j\big)}{\sum_{j=1}^J \mathcal{M}_j} ,
\]
where $\ell(\cdot,\cdot)$ is typically instantiated as the mean squared error (MSE) of $\ell(\hat{\mathrm{x}}_j, \mathrm{x}_j) = \left\| \hat{\mathrm{x}}_j - \mathrm{x}_j \right\|_2^2$. This formulation enforces the model to infer missing spatio-temporal spike patterns from partially observed context. 

\paragraph{Theoretical Justification:} 
\textbf{(1)} Context injection expands representation geometry. Spike tokens derived from neural recordings possess limited variability in firing patterns, which causes the projected embeddings to concentrate in a low-rank and crowded latent subspace. In contrast, metadata encoded by pretrained language models provides semantically rich and well-separated representations that are distributed more uniformly across the embedding space. By integrating metadata embeddings with spike tokens, UniBCI expands the occupied region of the latent space, yielding a more expressive and source-aware unified representation space.

\textbf{(2)} Linear attention is aligned with spike signal characteristics. Invasive spike signals are inherently sparse, locally structured, and temporally short-range within intervals. Fully connected self-attention introduces dense pairwise interactions across all temporal positions, which amplifies redundant correlations and optimization noise. Linear attention instead performs low-rank feature aggregation through associative accumulation, which matches the intrinsic low-rank structure of spike dependency. During an interval, this preserves dominant event patterns, providing a more suitable inductive bias for neural spike modeling.

\textbf{(3)} IAA learns a shared latent manifold. Neural population activity across different brain areas and temporal intervals is governed by shared latent neural dynamics. Adjacent intervals often belong to the same neural state evolution process, while neighboring brain regions exhibit correlated population activity driven by common task-related factors. IAA explicitly models this structure by combining interval-wise temporal modeling and area-wise locality-constrained interactions. This hierarchical interaction acts as manifold smoothing, aligning representations from different areas and intervals into a shared latent space. Detailed theoretical proof is provided in Appendix \ref{proof}.

\section{Experiments}
\subsection{Datasets and Data Processing}
We curated a diverse collection of datasets of neural spikes from non-human primates (NHPs), mice and humans to develop our model, as summarized in Table~\ref{tab:neural-datasets}. The pretraining corpus consists of large-scale recordings spanning multiple species, subjects, cortical and subcortical regions, and task paradigms. Specifically, we collect macaque datasets from M1-CO1, M1-CO2, Pac-Man and HPC-HG during center-out (CO) tasks, Pacman-game tracking and hidden goal navigation tasks. They encompass spike signals that span 189 hours covering brain regions of the primary motor cortex (M1), the premotor cortex (PMd), the dorsolateral prefrontal cortex (dlPFC), the anterior cingulate cortex (ACC) and the hippocampus (HPC). The pretraining corpus is further complemented by the LICK mouse dataset, which covers secondary motor cortex (M2), substantia nigra pars reticulata (SNR), and ventrolateral striatum (VLS) during licking behavior across 104 sessions. Downstream benchmarks contain both classification and regression tasks. The classification benchmarks include the M1-CO1, LICK, and PPC-FINGER~\cite{PPC} datasets, which involve center-out reaching, multi-session licking behavior, and human finger press movement decoding, respectively. We assessed the model’s generalization ability across different recording conditions and species. The regression benchmarks include the Perich ~\cite{perich2025long} dataset and the Neural Latent Benchmark suite~\cite{NLB}. Specifically, we evaluated cross-subject transfer on the Perich dataset in both center-out and random target (RT) reaching tasks. The MC-Maze~\cite{NLB} task investigates autonomous motor dynamics during continuous movement generation, while the Area2-Bump~\cite{NLB} task recording responses to externally driven sensory perturbations in the somatosensory cortex (S1). 

All datasets underwent standardized preprocessing, including spike binning, temporal alignment with behavioral kinematics (e.g., hand velocity or 2D coordinates), and sample segmentation via sliding windows or trial-based adaptive binning. More details refer to Appendix \ref{appendix_datasets}.
\begin{table}[t]
  \centering
  \caption{Pretraining and downstream datasets of invasive spike signals.}
  \label{tab:neural-datasets}
  \vspace{1em}
  \resizebox{\textwidth}{!}{
    \begin{tabular}{c|llccccc} 
      \toprule
      & \textbf{Dataset} & \textbf{Species} & \textbf{Brain Regions}& \textbf{Tasks} & \textbf{\# Indiv.} & \textbf{\# Sess.} &  \textbf{Data Scale}  \\
      \midrule
      \multirow{5}{*}{\rotatebox{90}{\scriptsize \textbf{Pretraining}}} 
      & M1-CO1& Macaque & M1      & CO     & 1 & 348 & 17.5 hours\\
      & M1-CO2     & Macaque & M1      & CO     & 1 & 29 & 17.8 hours\\
      & Pac-Man     & Macaque   & PMd, dlPFC, ACC       & Pacman & 1 & 121 & 109.5 hours\\
      & HPC-HG      & Macaque & HPC       & Hidden Goal      & 1 & 36 & 44.8 hours\\
      & LICK      & Mouse & M2, SNR, VLS       & Licking      & 11 & 104 &  74.1 hours\\
      \midrule
      \multirow{6}{*}{\rotatebox{90}{\scriptsize \textbf{Downstream}}} 
       & M1-CO1 (New Sess.)& Macaque & M1      & CO     & 1 & 24 & 1.3 hours\\
       & LICK (New Indiv.)      & Mouse & M2, SNR, VLS       & Licking      & 3 & 15 &  11.4 hours\\
       & PPC-FINGER~\cite{PPC}      & Human & PPC       & Finger Press      & 1 & 10 &  6.7 hours\\
      & Perich~\cite{perich2025long}& Macaque & M1, PMd & CO, RT& 1& 12& 2.5 hours\\
      & MC-Maze~\cite{NLB}& Macaque & M1, PMd      & Maze & 1 & 1 & 1.9 hours\\
      & Area2-Bump~\cite{NLB}& Macaque & S1      & Bump & 1 & 1& 0.6 hours \\
      \bottomrule
    \end{tabular}
  }
\end{table}
\subsection{Implementation and Settings}

\noindent\textbf{Model Implementation.}
During spike train normalization, the random channels are divided into 8 areas with the area size of 32, allowing the model to adapt to different electrode layouts or neural population structures. In the tokenization stage, we leverage a pretrained language model, MiniLM-L6-v2~\cite{minilmv2}, to encode textual metadata. The encoded textual features are projected into the same embedding space as neural signals via a linear projection layer. Next, the spatio-temporal tokens are partitioned into 10 intervals, each with 10 timesteps. We compose 4 interval-area attention blocks, each equipped with 8 heads and an embedding dimension of 64 for both interval linear attention (ILA) and area sliding window attention (ASWA) modules. We further set a window size of 10 in ASWA to model area-wise dependencies efficiently while maintaining computational tractability. 

\noindent\textbf{Pretraining Strategy.}
In pretraining, the masking ratio is set to 50\%. The model reconstructs the masked segments based on the visible tokens without behavioral labels. The training objective is defined as the mean squared error (MSE) between the reconstructed and ground-truth signals. We employ the AdamW optimizer with an initial learning rate of $5 \times 10^{-4}$ and apply a cosine annealing schedule for learning rate decay. The model is trained for 40 epochs with a batch size of 128. All experiments are conducted on 8 NVIDIA A100 GPUs.

\noindent\textbf{Evaluation Strategy.}
For downstream evaluation, the pretrained model is fine-tuned together with task-specific heads under a full-parameter training setting. The fine-tuning process uses a learning rate of $1 \times 10^{-4}$ and a batch size of 64. We utilized 80\% of each downstream dataset for training and reserving the remaining 20\% for testing. For M1-CO1, we adopt two settings: (1) Multi-day, where sessions from each days are split for joint training and testing; (2) Cross-day, where models are trained on a subset of days and tested on unseen days. For LICK, we train and test separately on each of the three individuals, and report the average performance across subjects. For all other datasets, evaluation is conducted within each session, with results averaged across sessions when applicable.

\noindent\textbf{Baselines \& Metrics.}
We compare the proposed method with several baselines on both classification and regression tasks using a comprehensive set of metrics.The traditional baselines include Wiener filters (WF)~\cite{wiener}, gated recurrent unit (GRU)~\cite{GRU}, a multilayer perceptron (MLP), and a variational autoencoder (VAE)~\cite{VAE}, where the VAE is pretrained on the same data as UniBCI. We fairly compare advanced neural representation learning models that POYO~\cite{poyo} is evaluated using its publicly released pretrained weights, while NDT1~\cite{ye2021ndt}, NDT2~\cite{NDT2}, and MtM~\cite{mtm} are evaluated using the pretrained models provided in corresponding articles.  For classification tasks, we report Balanced Accuracy and Weighted F1 score while for regression tasks, we utilize the coefficient of determination ($R^2$). More details refer to Appendix \ref{app_baselines}, \ref{app_metrics}.

\subsection{Downstream Experiment Results}
\begin{table}[t]
    \centering
    \caption{Performance comparison on classification and regression tasks. Values represent mean $\pm$ standard deviation where applicable. The \textit{Generalization} column indicates the transfer setting. \textbf{Bold} and \underline{underlined} values indicate the best and second-best performance, respectively.}
    \label{performance}
    \renewcommand{\arraystretch}{1.4}
    \setlength{\tabcolsep}{2.8pt} 
    \resizebox{\textwidth}{!}{
        \begin{tabular}{lllcccccccccc}
            \toprule
            \textbf{\textit{Generalization}} & \textbf{Dataset} & \textbf{Metrics} & \textbf{WF}~\cite{wiener} & \textbf{GRU}~\cite{GRU} & \textbf{MLP} & \textbf{VAE}~\cite{VAE} & \textbf{NoMAD}~\cite{nomad} & \textbf{NDT1}~\cite{ye2021ndt} & \textbf{NDT2}~\cite{NDT2} & \textbf{MtM}~\cite{mtm} & \textbf{POYO}~\cite{poyo} & \textbf{UniBCI (Ours)} \\
            \midrule
            
            \multicolumn{13}{c}{\textit{Classification Tasks}} \\ 
            \midrule
            
            \multirow{4}{*}{\textbf{\textit{New Session}}} & \multirow{2}{*}{M1-CO1 (Multi-day)} & B-Acc & 0.645 & 0.806 & 0.680 & 0.817 & 0.848 & 0.755 & 0.841 & 0.842 & \underline{0.875} & \textbf{0.895} \\
             & & F1-W & 0.649 & 0.808 & 0.680 & 0.820 & 0.850 & 0.756 & 0.841 & 0.844 & \underline{0.876} & \textbf{0.897} \\
             & \multirow{2}{*}{M1-CO1 (Cross-day)} & B-Acc & 0.623 & 0.739 & 0.666 & 0.758 & 0.821 & 0.816 & 0.829 & 0.825 & \underline{0.842} & \textbf{0.851} \\
             & & F1-W & 0.624 & 0.741 & 0.660 & 0.756 & 0.822 & 0.819 & 0.832 & 0.828 & \underline{0.839} & \textbf{0.851} \\
            
            \midrule
            \multirow{2}{*}{\textbf{\textit{New Subject}}} & \multirow{2}{*}{LICK} & B-Acc & 0.655$\pm$0.091 & 0.705$\pm$0.114 & 0.662$\pm$0.039 & 0.685$\pm$0.033 & 0.724$\pm$0.051 & 0.721$\pm$0.033 & 0.730$\pm$0.017 & \underline{0.734$\pm$0.007} & 0.730$\pm$0.053 & \textbf{0.744$\pm$0.019} \\
             & & F1-W & 0.792$\pm$0.052 & 0.781$\pm$0.055 & 0.851$\pm$0.024 & 0.843$\pm$0.017 & 0.855$\pm$0.026 & 0.858$\pm$0.030 & \textbf{0.862$\pm$0.010} & 0.861$\pm$0.015 & 0.857$\pm$0.021 & \underline{0.859$\pm$0.007} \\

            \midrule
            \multirow{2}{*}{\textbf{\textit{New Specie}}}& \multirow{2}{*}{PPC-FINGER} & B-Acc & 0.852$\pm$0.056 & 0.894$\pm$0.044 & 0.929$\pm$0.026 & 0.881$\pm$0.051 & 0.717$\pm$0.034 & 0.959$\pm$0.021 & \underline{0.961$\pm$0.022} & 0.934$\pm$0.026 & 0.940$\pm$0.033 & \textbf{0.967$\pm$0.020} \\
             & & F1-W & 0.824$\pm$0.053 & 0.894$\pm$0.043 & 0.927$\pm$0.027 & 0.884$\pm$0.045 & 0.719$\pm$0.037 & 0.958$\pm$0.018 & \underline{0.959$\pm$0.018} & 0.932$\pm$0.029 & 0.939$\pm$0.033 & \textbf{0.960$\pm$0.023} \\

             \midrule
            \multicolumn{3}{r}{\textbf{Avg.} B-Acc} & 0.694 & 0.786 & 0.734 & 0.785 & 0.778 & 0.813 & 0.840 & 0.834 & \underline{0.847} & \textbf{0.864} \\

            \midrule
            
            \multicolumn{13}{c}{\textit{Regression Tasks}} \\ 
            \midrule
            
            \textbf{\textit{New Subject}} & Perich T-CO & $R^2$ & 0.580$\pm$0.029 & 0.625$\pm$0.035 & 0.682$\pm$0.033 & 0.716$\pm$0.037 & 0.656$\pm$0.060 & 0.737$\pm$0.046 & 0.738$\pm$0.046 & 0.682$\pm$0.049 & \underline{0.753$\pm$0.024} & \textbf{0.757$\pm$0.021} \\
            
            \midrule
            \multirow{2}{*}{\textbf{\textit{New Task}}} & Perich T-RT & $R^2$ & 0.559$\pm$0.111 & 0.621$\pm$0.069 & 0.613$\pm$0.103 & 0.629$\pm$0.121 & 0.605$\pm$0.100 & 0.687$\pm$0.098 & 0.681$\pm$0.097 & 0.621$\pm$0.100 & \underline{0.692$\pm$0.111} & \textbf{0.702$\pm$0.100} \\
             & MC-Maze & $R^2$ & 0.728 & 0.773 & 0.804 & 0.797 & 0.795 & 0.882 & \underline{0.884} & 0.879 & 0.883 & \textbf{0.890} \\
            
            \midrule
            \textbf{\textit{New Region}} & Area2-Bump & $R^2$ & 0.724 & 0.814 & 0.778 & 0.797 & 0.793 & \textbf{0.902} & 0.888 & 0.874 & 0.882 & \underline{0.890} \\

            \midrule
            \multicolumn{3}{r}{\textbf{Avg. $R^2$}} & 0.648 & 0.708 & 0.719 & 0.735 & 0.712 & 0.802 & 0.798 & 0.764 & \underline{0.803} & \textbf{0.810} \\
            \bottomrule
        \end{tabular}
    }
\end{table}
We evaluate UniBCI across a wide range of downstream settings, with results reported in Table~\ref{performance}. Under the classification task, UniBCI achieves state-of-the-art performance in new session settings (M1-CO1), reaching a balanced accuracy of 0.895 on Multi-day and 0.851 on Cross-day, while also securing the best average accuracy (0.744) on new subject evaluation for LICK. On new species, it achieves a balanced accuracy of 0.967 on the human PPC-FINGER~\cite{PPC}, validating its cross-species transfer capability. For new tasks and brain regions, UniBCI yields an overall best average $R^2$ of 0.810 across Perich~\cite{perich2025long}, MC-Maze~\cite{NLB}, and Area2-Bump~\cite{NLB}, consistently surpassing strong baselines such as POYO~\cite{poyo} (0.803) and NDT1~\cite{ye2021ndt} (0.802). Overall, UniBCI consistently improves performance across all evaluation scenarios, demonstrating strong generalization of its pretrained neural representations, with effective transfer to new species, subjects, brain regions, and tasks. 

These gains are achieved with limited labeled data for fine-tuning, indicating that UniBCI’s context-aware embeddings and spatiotemporal hierarchical attention capture multi-scale neural dynamics and enhance robustness to domain shifts across heterogeneous recording conditions.

\subsection{Ablation Experiment Results}

We \begin{figure}[h]
    \centering
    \includegraphics[width=1\linewidth]{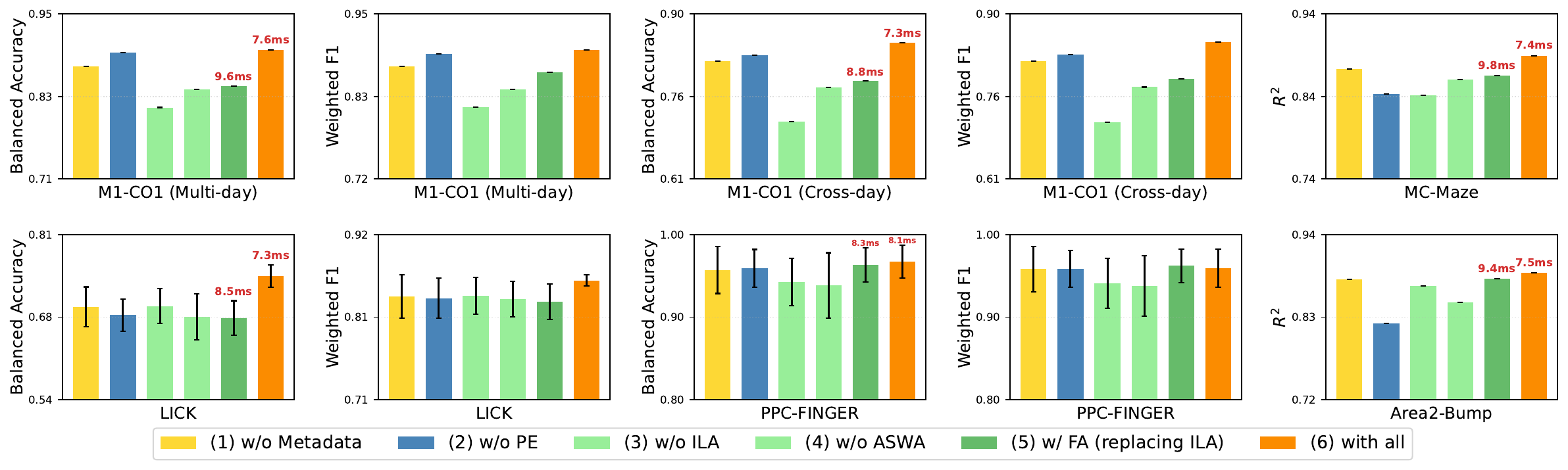}
    \caption{The experimental results of ablating modules. The red values in the subplot indicate the inference latency of Varient (5) and (6).}
    \label{ablation}
\end{figure} conduct ablation studies to evaluate the contribution of each component, as shown in Figure~\ref{ablation}. Removing metadata (Variant 1) leads to consistent performance drops across all datasets, with a 2.6\% decrease in balanced accuracy on M1-CO1 and a 1.8\% drop on MC-Maze, indicating that contextual information provides useful priors and helps mitigate domain shifts. Eliminating positional encoding (Variant 2) results in performance drops by 5.2\% and 7.5\% on MC-Maze and Area2-Bump respectively. This highlights the importance of explicit spatial and temporal order in modeling neural dynamics. Variants (3) and (4) evaluate the interval-area attention mechanism. Removing ILA causes a 9.1\% drop on M1-CO1, demonstrating its role in capturing fine-grained temporal patterns. Without the module of ASWA leads to consistent declines across datasets, including 6.3\% on M1-CO1 and 4.4\% on Area2-Bump, underscoring the importance of modeling global spatial dependencies. Furthermore, replacing ILA mechanism with full attention (Variant 5) causes consistent performance drops and increases latency, proving that ILA offers a more effective and efficient inductive bias for neural dependencies. The full model achieves the best performance across all benchmarks, confirming the complementary benefits of all components. These results show that metadata embeddings and hierarchical interval-area attention jointly enable effective multi-scale spatiotemporal representation learning. More replacement experiments of positional encodings and grouping mechanisms are provided in Appendix~\ref{add_res}.

            

\subsection{Sensitivity Analysis}
\begin{figure}[h]
    \centering
    \includegraphics[width=1\linewidth]{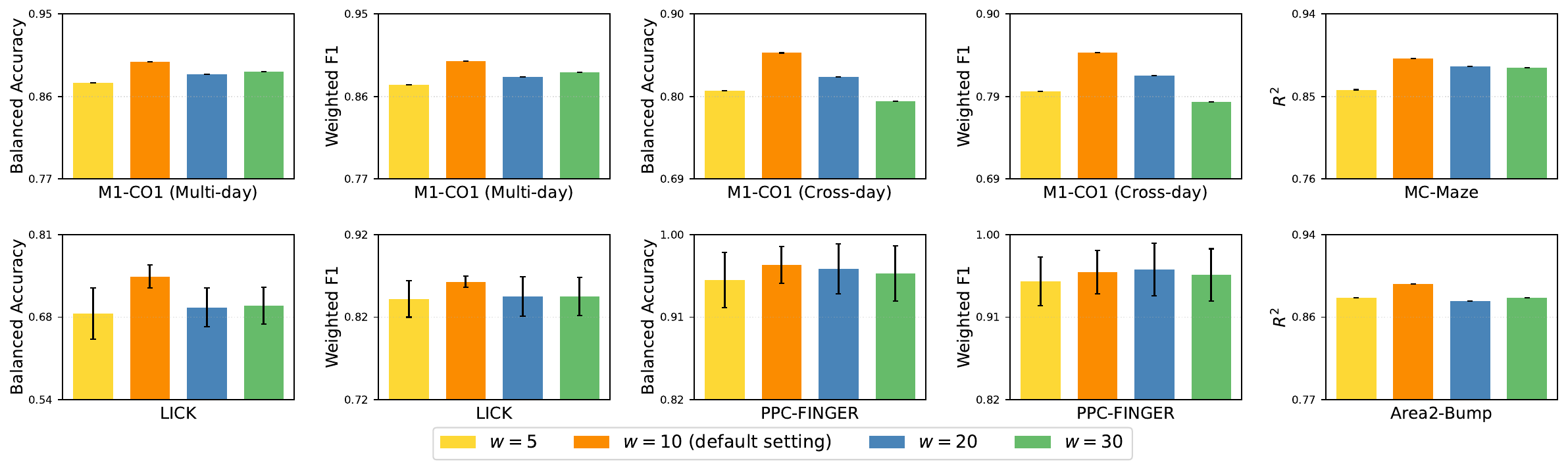}
    \caption{Sensitivity analysis of different sliding window sizes $w$ in ASWA.}
    \label{window_size}
\end{figure}
To further investigate the influence of the sliding window size $w$ in the Area-wise Sliding Window Attention (ASWA) mechanism, we conducted a sensitivity analysis by varying $w \in \{5, 10, 20, 30\}$. As shown in Figure~\ref{window_size}, the model's performance degrades when $w=5$. This occurs because a window size smaller than the number of areas ($A=8$) hinders the model's ability to establish necessary temporal dependencies across area boundaries, enabling an area in the current interval to attend to its corresponding features from the preceding interval. When $w$ is increased to 10 as our default setting, the model achieves the best performance, as this size is sufficient to bridge the inter-interval gap. Further increasing the window size to 20 or 30 does not yield significant gains, suggesting that neural dynamics are predominantly governed by local temporal dependencies. Therefore, a compact sliding window is sufficient for robust representation learning without introducing redundant computation or long-range noise. More sensitivity analysis of the number of areas and area size refer to Appendix \ref{add_res}.


\subsection{Efficiency and Complexity}
\begin{figure}[t]
    \centering
    \includegraphics[width=1\linewidth]{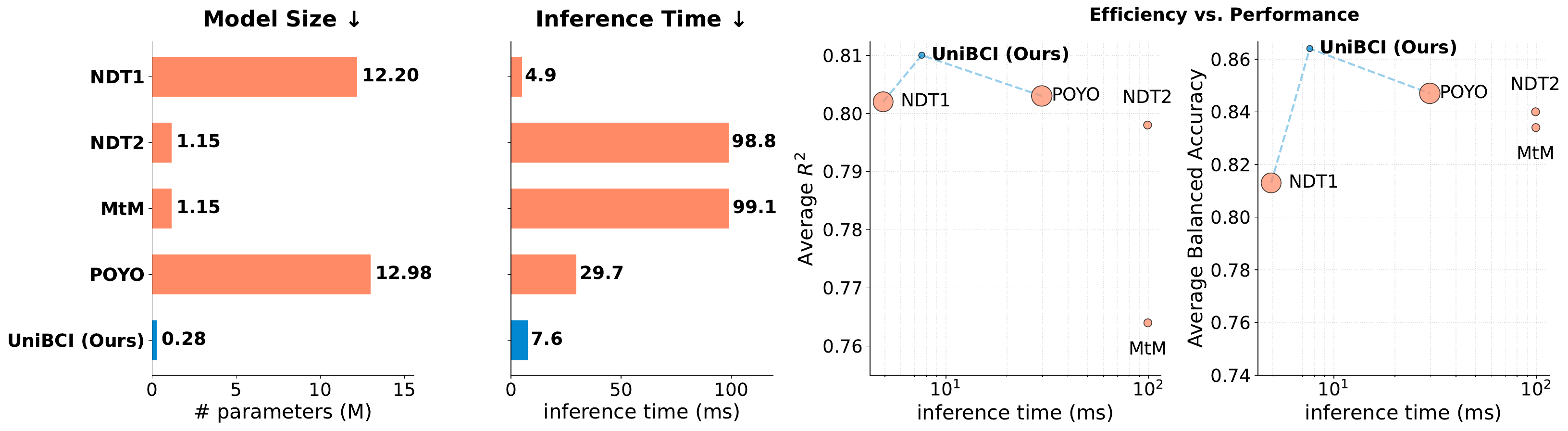}
    \caption{Comparison of model efficiency in terms of parameter count, inference latency and balanced performance. Note the x-axis is in log scale to better visualize the performance of low-latency models, and circle size represents the parameter scale.}
    \label{fig:efficiency}
\end{figure}

To evaluate computational efficiency, we report the number of trainable parameters of the model backbone and inference time per sample in Figure~\ref{fig:efficiency}. UniBCI incorporates a pretrained text encoder with 22.7M parameters, which is frozen during training and thus introduces no additional training cost. As shown in Figure~\ref{fig:efficiency}, UniBCI achieves strong model compactness, with only 0.28M trainable parameters, approximately 46 times smaller than NDT1~\cite{ye2021ndt} and POYO~\cite{poyo}. In terms of inference efficiency, NDT1 achieves the lowest latency (4.9 ms) but suffers from degraded accuracy and $R^2$ performance. In contrast, UniBCI maintains a competitive inference time of 7.6 ms while consistently outperforming all baselines, achieving both the highest average $R^2$ of 0.810 and balanced accuracy of 0.864. Overall, these results demonstrate that UniBCI effectively reduces redundant computation while preserving strong representational capacity under a highly parameter-efficient design, making it suitable for real-time invasive brain-computer interfaces.
\section{Conclusion}
We present UniBCI, a unified neural foundation model for invasive spike data that addresses key challenges in neural decoding, including data heterogeneity with limited scale, cross-domain distribution shift and spatiotemporal complexity of invasive signals. By integrating context-conditioned tokenization, the proposed framework enables scalable and generalizable representation learning across diverse neural datasets. We introduce a hierarchical interval-area attention module to capture fine-grained features of spike signals across intervals and areas. In pretraining, self-supervised masked reconstruction requires no behavioral labels, thereby optimizing UniBCI efficiently. Comprehensive experiments demonstrate that UniBCI consistently outperforms existing methods on both classification and regression benchmarks, suggesting strong generalization across species, subjects, sessions and task paradigms. Moreover, the model maintains high efficiency with a compact architecture, making it suitable for real-time BCI applications where both accuracy and latency are critical. Future work will explore more adaptive tokenization schemes based on the biological spatial structure and multimodal integration. Overall, this work highlights the potential of foundation model paradigms in neuroscience and provides a step toward unified, general-purpose modeling of neural population dynamics.
\bibliography{ref}
\bibliographystyle{unsrt}

\newpage
\appendix
\section{RELATED WORK}

\subsection{Neural Decoding Methods}
Neural decoding aims to infer behavioral or cognitive variables from neural activity. Early approaches were primarily based on linear statistical models and dynamical systems, such as Kalman filters~\cite{Kalman}, Wiener filters~\cite{wiener}, and generalized linear models (GLMs). While effective at capturing low-dimensional latent structures, these methods rely on linearity and stationarity assumptions, limiting their ability to model complex nonlinear neural dynamics. With the rise of deep learning, architectures including recurrent neural networks (RNNs)~\cite{RNN}, convolutional neural networks (CNNs)~\cite{CNN}, Transformers~\cite{Transformers}, and state space models (SSMs)~\cite{SSM} have been widely adopted for neural signal modeling, improving performance in tasks such as motor decoding and behavior prediction. These models better capture nonlinear temporal dependencies, but are typically restricted to single-session or single-task settings, and still suffer from limited generalization across sessions, subjects, and experimental conditions. To improve cross-session stability, alignment-based methods have been proposed. For instance, CycleGAN~\cite{cycleGAN} and NoMAD~\cite{nomad} train a decoder on a reference session (e.g., day one) and subsequently learn mappings to align neural activity from later sessions. AlignNet~\cite{hong2026bidirectional} further integrates spiking neural networks (SNNs), artificial neural networks (ANNs), and contrastive learning to align neural signals and behavioral semantics into a shared latent space, enabling robust cross-day bidirectional decoding and simulation. Despite their effectiveness under controlled settings, these methods still exhibit notable generalization gaps when applied to unseen subjects or tasks.

\subsection{Foundation Models for Invasive Signals}
The emergence of large-scale self-supervised foundation models is reshaping machine learning, demonstrating strong cross-task transferability in domains such as natural language processing~\cite{gpt1, gpt2, flan} and computer vision~\cite{clip, sam}. This paradigm is increasingly being extended to neuroscience. In neural decoding, recent works explore large-scale pretraining strategies. On the self-supervised side, NDT1~\cite{ye2021ndt} introduces masked modeling for neural spike sequence reconstruction using a Transformer, enabling parallel (non-recurrent) temporal modeling via masked time segments. NDT2~\cite{NDT2} extends this framework to a spatiotemporal Transformer and improves generalization through cross-session and cross-subject pretraining. MtM~\cite{mtm} further proposes multi-level masking strategies (across time, neurons, and brain regions) to learn more general neural representations. Under behavior-supervised settings, POYO~\cite{poyo} leverages Perceiver IO–based cross-attention to tokenize spike events, enabling large-scale multi-session training without explicit neuron alignment; its extension, POYO+~\cite{poyo+}, enhances adaptability across cell types and brain regions. NDT3~\cite{ndt3} scales autoregressive Transformers to hundreds of heterogeneous datasets, achieving strong zero-shot and few-shot generalization. POSSM~\cite{POSSM} combines a cross-attention encoder with a recurrent SSM backbone, improving inference efficiency while maintaining competitive performance. NEDS~\cite{neds} introduces a multi-task masking strategy, in which masking is alternated across neural, behavioral, and cross-modal modalities during training, enabling the model to jointly learn representations that are consistent across heterogeneous supervision signals. Nevertheless, existing approaches are still largely confined to single-species or limited experimental domains, lacking a unified framework that generalizes across brain regions, species, and tasks.

\subsection{Spatiotemporal Modeling of Neural Dynamics}
Neural signals exhibit rich multi-scale spatiotemporal dependencies, making their effective modeling a central challenge in neural decoding. Traditional approaches, such as RNNs~\cite{RNN} and SSMs~\cite{SSM}, are capable of modeling temporal dependencies but often struggle with long-range interactions. Transformers, with self-attention mechanisms, enable direct modeling of dependencies across arbitrary time steps, substantially improving long-sequence modeling. In non-invasive neural signals (e.g., EEG and MEG), Transformer-based spatiotemporal models have been widely explored. For example, CBraMod~\cite{wang2025cbramod} introduces a criss-cross attention mechanism that decouples temporal and spatial modeling to address signal heterogeneity, while CSBrain~\cite{zhou2025csbrain} proposes a cross-scale spatiotemporal framework with hierarchical tokenization and structured sparse attention to capture features ranging from local activations to slow oscillatory dynamics. For invasive neural signals, NDT2~\cite{NDT2} adopts a ViT-style~\cite{vit} spatiotemporal patch tokenization, where recorded neurons (channels) are spatially arranged and grouped into tokens for unified modeling of neural dynamics. However, this design incurs substantial computational overhead. Building upon this paradigm, our tokenization strategy uniformly groups neurons and further partitions them into local blocks. We apply interval linear attention alongside area sliding window attention, reducing computational complexity while preserving the ability to capture both local and global spatiotemporal dependencies.

\section{MODAL DETAILS AND EXPERIMENTAL SETTINGS}
\label{app_exp}
\subsection{Model Details and Pretraining Settings}
To achieve cross-dataset alignment and generalizable neural representation learning, we design a transformer-based pretraining model, with detailed configurations reported in Table~\ref{model_pre}. The architecture adopts a 64-dimensional embedding space and consists of 4 interval-area attention blocks, each equipped with 8 attention heads. At the input stage, the raw neural signals are segmented into 10 intervals, each with a length of 10 time steps. We further integrate a pretrained MiniLM-L6-v2 as the text encoder, enabling natural language guidance to facilitate the learning of shared structures across heterogeneous datasets. During pretraining, the model is trained on 8 A100 GPUs with a batch size of 128 for 40 epochs. We adopt the AdamW optimizer with an initial learning rate of $5 \times 10^{-4}$, together with a cosine annealing schedule for dynamic learning rate decay. A masking ratio of 0.5 is applied during training, encouraging the model to learn robust neural representations via a reconstruction objective under partial observation.
\begin{table}[!h]
\centering
\caption{Model architecture and pretraining hyperparameters.}
\label{tab:model_config}
\begin{tabular}{lcc}
\toprule
\textbf{Category} & \textbf{Hyperparameters} & \textbf{Settings} \\ 
\midrule
\multirow{4}{*}{Architecture} 
& Embedding Dimension& 64 \\
& Number of Layers & 4 \\
& Attention Heads & 8 \\
& Sliding Attention Window Size & 10 \\
\midrule
\multirow{3}{*}{Tokenization} 
& Interval Size & 10 \\
& Area Size & 32 \\
& Area Number & 8 \\
\midrule
\multirow{2}{*}{Language} 
& Text Encoder & MiniLM-L6-v2 \\
& Text Encoding & Enabled \\
\midrule
\multirow{9}{*}{Pre-training} 
& Epochs & 40 \\
& Batch Size & 128 \\
& Masking Ratio & 0.5 \\
& Dropout & 0.1 \\
& Optimizer & AdamW \\
& Learning Rate & 5e-4 \\
& Weight Decay & 5e-2 \\
& Scheduler & CosineAnnealingLR \\
& Minimal Learning Rate & 1e-5\\
\bottomrule
\end{tabular}
\label{model_pre}
\end{table}

\subsection{Downstream Experimental Settings}
During downstream finetuning, we load the pretrained weights of UniBCI and replace the original reconstruction head with a task-specific head composed of a multi-layer perceptron (MLP). The learned representations are flattened and fed into this head for downstream classification or regression tasks. We then perform full-parameter finetuning on the target datasets while freezing the text encoder. For classification tasks, we use the cross-entropy loss, whereas for regression tasks, we adopt the mean squared error (MSE) loss.

\section{THEORETICAL PROOF}
\label{proof}



\subsection{Metadata Injection Expands the Semantic Space of Spike Tokens}
Let the spike embedding after projection be
\begin{equation}
X_s \in \mathbb{R}^{n \times d},
\end{equation}
where $d$ is the hidden dimension. Due to the sparsity and binary event-driven nature of invasive spike trains, the intrinsic dimensionality of $X_s$ is much smaller than the ambient dimension:
\begin{equation}
\mathrm{rank}(X_s) = r_s \ll d.
\end{equation}
Equivalently, its covariance matrix
\begin{equation}
\Sigma_s = \frac{1}{n} X_s^\top X_s
\end{equation}
is low-rank, and its differential entropy satisfies
\begin{equation}
H(X_s) \leq \frac{1}{2}\log \det (2\pi e \Sigma_s).
\end{equation}

Since most eigenvalues of $\Sigma_s$ are close to zero, we have
\begin{equation}
\det(\Sigma_s) \rightarrow 0,
\end{equation}
which implies latent crowding and weak inter-source separability.

Let metadata be encoded by a pretrained language model as
\begin{equation}
X_m \in \mathbb{R}^{n \times d},
\end{equation}
whose semantic representation spans a high-rank manifold:
\begin{equation}
\mathrm{rank}(X_m) = r_m, \quad r_m \approx d.
\end{equation}

The final token representation is
\begin{equation}
Z = X_s + X_m.
\end{equation}

Assuming weak cross-correlation between spike features and metadata semantics,
\begin{equation}
\Sigma_Z = \Sigma_s + \Sigma_m.
\end{equation}

By Minkowski determinant inequality,
\begin{equation}
\det(\Sigma_Z)^{1/d}
\geq
\det(\Sigma_s)^{1/d}
+
\det(\Sigma_m)^{1/d},
\end{equation}
thus
\begin{equation}
\det(\Sigma_Z) > \det(\Sigma_s), \quad H(Z) > H(X_s).
\end{equation}

For source discrimination, consider two samples $i,j$:
\begin{equation}
\|Z_i - Z_j\|_2^2
=
\|X_i - X_j\|_2^2
+
\|M_i - M_j\|_2^2
+
2\langle X_i - X_j, M_i - M_j\rangle.
\end{equation}

Since semantic metadata embeddings are approximately orthogonal,
\begin{equation}
\mathbb{E}\big[\langle X_i - X_j, M_i - M_j\rangle\big] \approx 0,
\end{equation}
we obtain
\begin{equation}
\mathbb{E}\|Z_i - Z_j\|_2^2 > \mathbb{E}\|X_i - X_j\|_2^2.
\end{equation}

\subsection{Advantage of Linear Attention for Spike Signals}

Given an input $H \in \mathbb{R}^{t \times d}$ within an interval, standard attention computes:
\begin{equation}
\mathrm{Attn}(H) = \mathrm{softmax}(QK^\top)V,
\end{equation}
with $Q = HW_q, \; K = HW_k, \; V = HW_v$.

Linear attention reformulates this as:
\begin{equation}
\mathrm{Attn}(H) = Q (K^\top V).
\end{equation}

For spike signals, where activations are sparse and temporally localized, we model:
\begin{equation}
H = S + \epsilon,
\end{equation}
where $S$ is a sparse signal component and $\epsilon$ is noise.

Fully attention introduces pairwise interactions:
\begin{equation}
QK^\top \sim SS^\top + S\epsilon^\top + \epsilon S^\top + \epsilon\epsilon^\top,
\end{equation}
which amplifies noise via cross terms.

In contrast, linear attention aggregates as:
\begin{equation}
K^\top V \sim S^\top S + \text{lower-order noise terms},
\end{equation}
which suppresses noise due to summation before interaction with queries. Therefore, linear attention provides a more stable estimator of co-activation patterns, making it better suited for spike data.

\subsection{Shared Manifold Learning via Interval-Area Attention}

Let $H_{i,a} \in \mathbb{R}^{t \times d}$ denote the representation of interval $i$ and area $a$. We assume neural activity is generated from a shared latent dynamical state $z_i \in \mathbb{R}^{d_z}$:
\begin{equation}
H_{i,a} = f_a(z_i) + \epsilon_{i,a},
\end{equation}
where $f_a: \mathbb{R}^{d_z} \rightarrow \mathbb{R}^{t \times d}$ is an area-specific smooth mapping and $\epsilon_{i,a}$ is zero-mean noise with bounded variance.

\paragraph{Interval-wise Encoding.}
The interval attention module applies a transformation $g(\cdot)$ to extract temporal features:
\begin{equation}
\tilde{O}_{i,a} = g(H_{i,a}).
\end{equation}
Under the assumption that $g(\cdot)$ is Lipschitz continuous and approximately preserves local structure, we have:
\begin{equation}
\tilde{O}_{i,a} \approx \tilde{f}_a(z_i) + \tilde{\epsilon}_{i,a},
\end{equation}
where $\tilde{f}_a$ is a transformed mapping and $\tilde{\epsilon}_{i,a}$ remains bounded.

\paragraph{Area-wise Alignment via Attention.}
After temporal pooling, denote $\bar{H}_{i,a} \in \mathbb{R}^d$ as the aggregated representation. The area-wise attention computes:
\begin{equation}
O_{i,a} = \sum_{(j,b) \in \mathcal{N}(i,a)} \alpha_{(i,a),(j,b)} \, \bar{H}_{j,b},
\end{equation}
where $\mathcal{N}(i,a)$ is the sliding window neighborhood and $\alpha$ are attention weights.

Substituting the generative model:
\begin{equation}
O_{i,a} \approx \sum_{(j,b)} \alpha_{(i,a),(j,b)} \, f_b(z_j).
\end{equation}

If the latent dynamics are smooth across nearby intervals, i.e.,
\begin{equation}
\| z_i - z_j \| \le \delta \quad \text{for } (j,b) \in \mathcal{N}(i,a),
\end{equation}
then by smoothness of $f_b$:
\begin{equation}
f_b(z_j) \approx f_b(z_i).
\end{equation}

Thus, the aggregation simplifies to:
\begin{equation}
O_{i,a} \approx \sum_{b} \beta_{a,b} f_b(z_i),
\end{equation}
which is a weighted combination of mappings evaluated at the same latent state $z_i$.

\paragraph{Emergence of a Shared Manifold.}
Define a unified representation:
\begin{equation}
\phi(z_i) = \sum_{b} \beta_{a,b} f_b(z_i).
\end{equation}
Then:
\begin{equation}
O_{i,a} \approx \phi(z_i),
\end{equation}
which is independent of the specific area $a$ up to weighting coefficients.

Therefore, all representations $\{O_{i,a}\}_a$ lie near a shared manifold:
\begin{equation}
\mathcal{M} = \{ \phi(z) \mid z \in \mathbb{R}^{d_z} \}.
\end{equation}

\paragraph{Stacked Refinement.}
With multiple layers, the residual updates:
\begin{equation}
H^{\ell+1} = \mathcal{F}(H^{\ell}) + H^{\ell}
\end{equation}
iteratively reduce the variance of $\epsilon_{i,a}$ and enforce consistency across areas. This can be interpreted as a contraction mapping toward $\mathcal{M}$, leading to progressively aligned representations.

\paragraph{Conclusion.}
Interval-Area Attention acts as a structured operator that (i) preserves local temporal dynamics, and (ii) enforces cross-area consistency through attention-based aggregation. Under mild smoothness assumptions, this results in learning a shared latent manifold that explains neural activity across both spatial areas and temporal intervals.

\section{COMPUTATIONAL COMPLEXITY ANALYSIS}
\label{theory}
We analyze the computational complexity of the proposed Interval-Area Attention (IAA), which consists of interval linear attention (ILA) and area-wise sliding window attention (ASWA). Let the input feature be ${H} \in \mathbb{R}^{N \times A \times t \times d}$, where $N$ denotes the number of temporal intervals, $A$ the number of grouped areas, $t$ the token length within each interval, and $d$ the embedding dimension.

\subsection{Interval Linear Attention}

For each interval-area pair ${H}_{i,a} \in \mathbb{R}^{t \times d}$, ILA applies linear attention independently:
\begin{equation}
\tilde{O}_{i,a} = H_{i,a}W_q \cdot \left((H_{i,a}W_k)^\top H_{i,a}W_v\right).
\end{equation}
Unlike standard self-attention with quadratic complexity $\mathcal{O}(t^2d)$, linear attention first computes the key-value interaction $(K^\top V)$ with complexity $\mathcal{O}(td^2)$, followed by query projection with complexity $\mathcal{O}(td^2)$. Therefore, the complexity for one interval-area pair is $\mathcal{O}(td^2)$. Since computation is performed over all $N \times A$ interval-area pairs, the total complexity of ILA is $\mathcal{O}(NAtd^2)$.

\subsection{Area-wise Sliding Window Attention}

After temporal pooling, the token sequence is reduced to $S = N \times A$, and the feature of spike tokens denotes as $\tilde{H} \in \mathbb{R}^{S \times d}$. Standard global self-attention across all interval-area tokens requires:
\begin{equation}
\mathcal{O}(S^2 d).
\end{equation}
To avoid quadratic growth, ASWA restricts each query to a local sliding window of size $w$. Thus, each token only attends to its neighboring $w$ tokens, and the total complexity is reduced to $\mathcal{O}(Swd)$. Since $w$ is constant and independent of sequence length, the complexity becomes approximately linear with respect to $S$ : $\mathcal{O}(Sd)$.
\section{CODE ACCESS}
\label{code}
The code of this paper is accessible at: \url{https://anonymous.4open.science/r/UniBCI-C805}.

\section{DATASETS}\label{appendix_datasets}
\subsection{Pretraining Datasets}
\subsubsection{M1-CO1 Dataset}
The M1-CO1 dataset records neural activity from a macaque performing a standard eight-direction center-out reaching task. In this task, the subject moves a cursor from the center toward one of eight peripheral targets based on visual cues. Unlike continuous control tasks, the labels are discrete movement directions, represented as categorical classes ranging from 0 to 7. Neural signals were recorded from the primary motor cortex (M1).

\textbf{Neural and Behavioral Signals:}
The dataset is stored in HDF5 format and contains sorted spike trains. Each file corresponds to a single trial. The recordings include up to 232 neural units, providing high-resolution population activity. The behavioral label for each trial is a single discrete direction index.

\textbf{Preprocessing:}
To standardize inputs for the model, variable-length spike trains are resampled using a temporal binning strategy. Specifically, an adaptive binning approach is applied to divide each trial into 100 equal segments along the temporal axis. Spike events within each bin are aggregated to form spike count sequences, resulting in a fixed-length representation of 100 time steps. During pretraining, a total of 348 sessions are used.

\subsubsection{M1-CO2 Dataset}

The M1-CO2 dataset records neural and behavioral activity from a macaque performing an eight-direction center-out reaching task in a whack-a-mole paradigm using a robotic interface. Neural signals were recorded from the contralateral primary motor cortex (M1) via a 64-channel electrode array. The dataset consists of 29 recording days collected in April 2025, comprising a total of 32,083 valid trials.

\textbf{Neural and Behavioral Signals:}
Raw neural signals were recorded at a sampling rate of 30 kHz across 64 channels. Behavioral signals, consisting of two-dimensional joystick voltage trajectories, were simultaneously recorded at 1 kHz.

\textbf{Preprocessing:}
Spike detection was first applied to convert continuous neural voltage signals into discrete spike events, forming a two-dimensional spike matrix. To ensure temporal alignment, behavioral signals were linearly interpolated to 30 kHz and synchronized with neural recordings. Both neural and behavioral data were then downsampled to 50 Hz using temporal binning. Gaussian smoothing was applied to the behavioral trajectories to improve the signal-to-noise ratio. Finally, the data were segmented into fixed-length samples of 2 seconds, corresponding to 100 time steps, forming 64-dimensional spike sequence inputs paired with 2D trajectory decoding targets.

\subsubsection{Pac-Man Dataset}

The Pac-Man dataset records neural and behavioral activity from a macaque performing a Pacman-based continuous tracking task. The subject controls an on-screen character using a joystick to complete the task. Neural signals were recorded from multiple brain regions involved in motor planning and execution, primarily the premotor cortex (PMd), as well as the dorsolateral prefrontal cortex (dlPFC) and anterior cingulate cortex (ACC). The dataset consists of continuous recording sessions spanning multiple days. The sampling rate is 60 Hz.

\textbf{Neural and Behavioral Signals:}
The dataset contains sorted single-unit spike trains and synchronized behavioral trajectories. Due to variations in recording conditions across days, the number and anatomical locations of identified neurons vary dynamically. Behavioral data consist of two-dimensional position coordinates (X, Y) of the Pacman agent during task execution.

\textbf{Preprocessing:}
Neural spike trains and behavioral trajectories are first temporally aligned across brain regions. To smooth neural firing rates and extract temporal features, a 100 ms binning strategy is applied, and behavioral positions are averaged within each bin. A sliding window approach is then used to segment continuous recordings into samples, with each window spanning 10 seconds (100 time bins). Data augmentation is performed using a 2-second stride (20 time bins). During pretraining, recordings spanning 121 days are used.

\subsubsection{HPC-HG Dataset}

The HPC-HG dataset records neural activity from a macaque (Monkey P) performing a free-moving foraging task in a large 3D arena (3.5 m $\times$ 3.5 m $\times$ 2.8 m). The subject is required to search for hidden reward locations. Once the animal enters a reward zone, it must visit a randomly illuminated touchscreen located around the arena to obtain food rewards. The reward location is shifted after 50 successful retrievals.

\textbf{Neural and Behavioral Signals:}
Neural signals were recorded from the hippocampus (HPC), which is closely associated with spatial representation and memory. The dataset contains spike trains from multiple sorted single units, sampled at 100 Hz (10 ms resolution). Due to cross-day variability, the number of identified units varies across sessions. Behavioral data consist of synchronized two-dimensional spatial coordinates (X, Y) of the animal within the arena.

\textbf{Preprocessing:}
Invalid behavioral samples (NaN values) and their corresponding neural segments are removed using a synchronization mask to ensure precise alignment between hippocampal activity and spatial trajectories. Spike trains are then binned using a 100 ms window, and spike counts are accumulated within each bin. The corresponding spatial coordinates are averaged over the same temporal bins. A sliding window approach is applied to segment the continuous recordings into samples, each containing 10 seconds of data (100 time bins), with a stride of 2 seconds for data augmentation. During pretraining, recordings spanning 36 days are used.

\subsubsection{LICK Dataset}

The LICK dataset records neural activity and licking behavior in water-deprived mice performing a periodic water-reward task. The dataset includes 14 adult male mice with a recording duration of 119 days and a total of 28,573 trials. The animals are head-fixed and trained to perform a fixed-interval licking task in which a sucrose reward of 2 to 3 $\mu$L is delivered every 10 seconds. Neural signals were recorded from three brain regions including secondary motor cortex (M2) in 5 mice, substantia nigra pars reticulata (SNR), and ventrolateral striatum (VLS) in 9 mice.

\textbf{Neural and Behavioral Signals:}
Neural activity was recorded using a 32-channel Cerebus system at 30 kHz sampling rate. Behavioral signals were simultaneously recorded using electrical or infrared sensors and converted into a binary sequence, where 0 indicates no licking and 1 indicates licking. The released dataset is downsampled to 2,000 Hz to ensure precise temporal alignment between neural spikes and behavioral events.

\textbf{Preprocessing:}
The preprocessing pipeline applies both binning and sliding window segmentation. First, the data are downsampled from 2,000 Hz to 100 Hz by accumulating spike counts within each bin. Behavioral labels are aggregated using a max operation, such that any licking event within a bin is assigned a label of 1, ensuring precise alignment between neural dynamics and discrete behavioral events. A sliding window strategy is then used to segment the continuous recordings, with a fixed window length of 100 time steps and a stride of 95. During pretraining, recordings from 11 mice across 104 days are used, forming fixed-dimensional neural spike sequences paired with binary behavioral prediction targets.

\subsection{Downstream Datasets}
\subsubsection{M1-CO1 Dataset}
For downstream evaluation, we use 24 sessions spanning 10 days that were not included in pretraining. The task is an eight-class classification of movement directions. We evaluate M1-CO1 under two protocols: (1) \textit{Multi-day}, where all sessions across the 10 recording days are pooled and randomly split at the session level with an 80\%/20\% ratio for training and testing, ensuring that both sets contain trials from all directions; and (2) \textit{Cross-day}, where data from 8 days are used for training and the remaining 2 days are held out for testing to assess temporal generalization across recording days. The neural and behavioral signals, as well as preprocessing steps, follow the same procedures described in the pretraining section.

\subsubsection{LICK Dataset}

For downstream evaluation, we use data from three mice that are not seen during pretraining, with one mouse selected from each brain region. For each mouse, the data are randomly split into training and testing sets with an 80\% to 20\% ratio. The task is formulated as a binary classification problem of licking behavior prediction. The final performance is reported by averaging the results across the three mice.

\subsubsection{PPC-FINGER Dataset}

The PPC-FINGER dataset\footnote{\url{https://doi.org/10.48324/dandi.000147/0.221122.2256}} contains invasive neural recordings from a single tetraplegic participant performing a 6-class BCI finger press task. Neural activity was recorded from the left posterior parietal cortex (PPC, PC-IP region) using a 96-channel Utah array (Blackrock Microsystems). In each trial, one of six finger cues was presented at a random screen location, and the participant attempted to press the corresponding finger following visual instruction. Each trial lasts approximately 1.5 s after cue onset, capturing neural activity during motor intention and execution planning.

\textbf{Neural and Behavioral Signals:}
The dataset provides sorted spike trains from all channels, together with trial-level annotations including 6-class finger labels, cue locations, and precise temporal boundaries (start/end of each trial).

\textbf{Preprocessing:}
Spike times are extracted from NWB files and aligned to trial events using the recorded sampling rate. Signals are binned into fixed-length temporal windows to form spike count tensors of shape $[\mathrm{Trials}, \mathrm{Time}, \mathrm{Channels}]$. The data are then split into training and testing sets with an 80\%/20\% ratio. The task is formulated as a 6-class classification problem for decoding finger movement intentions from PPC activity.

\subsubsection{Perich Dataset}

The Perich dataset\footnote{\url{https://doi.org/10.48324/dandi.000688/0.250122.1735}} consists of 117 recording sessions collected from four non-human primates, denoted as Monkey C, M, J, and T, covering a range of motor control and planning tasks. The subjects are seated and control a cursor on a two-dimensional screen using a joystick. The tasks include two paradigms. The center-out task involves reaching from a central position to one of eight targets arranged on a circle with a radius of 8 cm. Each trial includes a hold period, a delay until a go cue, and a subsequent reaching movement. The random target task involves continuously moving between randomly placed targets instead of fixed circular positions.

\textbf{Neural and Behavioral Signals:}
Neural activity was recorded using chronic multi-electrode arrays implanted in the primary motor cortex (M1) and dorsal premotor cortex (PMd). The raw data were acquired using the Blackrock Cerebus system and processed with single-unit spike sorting. Channels with high crosstalk and units with excessive coincident spikes were removed to ensure neuron independence.

\textbf{Preprocessing:}
For downstream evaluation, we use 12 sessions from Monkey T, including 6 center-out sessions and 6 random target sessions, which are completely held out from pretraining. Data preprocessing, binning, and dataset splitting are performed using the brainsets toolkit provided in~\cite{poyo}, ensuring a standardized and reproducible pipeline. The data are randomly split into training and testing sets with an 80\% to 20\% ratio. Neural signals are binned with a temporal resolution of 10 ms to obtain spike count sequences. The data are then segmented into fixed-length sequences of 1.0 second, corresponding to 100 time steps, forming standardized samples for motor decoding tasks.

\subsubsection{MC-Maze Dataset}
The MC-Maze dataset\footnote{\url{https://doi.org/10.48324/dandi.000128/0.220113.0400}} is derived from the Neural Latent Benchmark and records motor cortical activity from a macaque performing an obstacle-constrained maze reaching task. The subject performs a delayed center-out reaching task with obstacle constraints. In this task, the macaque must navigate around obstacles on the screen to reach a target. Due to varying obstacle configurations, the task generates highly stereotyped linear and curved movement trajectories. The experiment simultaneously records cursor position, eye gaze, and two-dimensional hand position and velocity.

\textbf{Neural and Behavioral Signals:}
Neural activity was recorded from the dorsal premotor cortex (PMd) and primary motor cortex (M1). Prior studies have shown that neural dynamics in M1 and PMd exhibit strong predictability around movement onset, reflecting pronounced autonomous dynamics.

\textbf{Preprocessing:}
The dataset is standardized for dynamical modeling. Neural signals are processed using a temporal resolution of 7 ms. We follow the data acquisition and preprocessing pipeline provided in~\cite{song2025langevin} for dataset downloading and processing. The dataset uses the default train-test partitioning released with the original data, corresponding to an approximate 75\%/25\% split. The behavioral targets are the two-dimensional hand velocities $(v_x, v_y)$, and the task is formulated as a regression problem to evaluate the model's ability to extract stable neural representations under limited data conditions.

\subsubsection{Area2-Bump Dataset}
The Area2-Bump dataset\footnote{\url{https://doi.org/10.48324/dandi.000127/0.220113.0359}} is derived from the Neural Latent Benchmark and records somatosensory cortical activity from a macaque performing a center-out reaching task with mechanical perturbations. The task follows a standard center-out reaching paradigm, with additional mechanical perturbations applied in a subset of trials. Specifically, the subject's arm is subjected to random force perturbations before movement onset, requiring the animal to adjust its trajectory to reach the target. The experiment records rich multimodal kinematic signals, including hand position, velocity, acceleration, interaction forces between the hand and joystick, as well as muscle lengths, muscle velocities, joint angles, and joint velocities.

\textbf{Neural and Behavioral Signals:}
Neural activity was recorded from the somatosensory cortex (S1), specifically Brodmann area 2. Unlike motor cortical regions that exhibit strong autonomous dynamics, neural responses in this area are strongly driven by external perturbations and encode detailed whole-arm kinematics.

\textbf{Preprocessing:}
The dataset is standardized for dynamical modeling. Neural signals are processed with a temporal resolution of 6 ms. We follow the data acquisition and preprocessing pipeline provided in~\cite{song2025langevin} for dataset downloading and processing. The dataset uses the default train-test partitioning released with the original data, corresponding to an approximate 75\%/25\% split. Behavioral targets are the hand velocity signals, and the task is formulated as a regression problem to evaluate the model's ability to learn neural representations under complex non-autonomous sensory inputs.

\section{BASELINES IMPLEMENTATION}
\label{app_baselines}
\noindent{\textbf{WF~\cite{wiener}:}}
We adopt the Wiener filter (WF) as a traditional linear decoding baseline. WF models the mapping from neural spike activity to behavioral outputs using a linear regression framework with temporal filtering. It serves as a strong classical benchmark for neural decoding tasks and is trained directly on each downstream task without any pretraining. For fair comparison, the same data splits and evaluation protocols are used as in all other methods.

\noindent{\textbf{GRU~\cite{GRU}:}}
As a sequential modeling baseline, we implement a gated recurrent unit (GRU) decoder to capture temporal dependencies in neural spike sequences. The input spike tensor is fed into a multi-layer GRU, followed by a linear prediction head for classification or regression tasks. The GRU is trained end-to-end directly on downstream tasks without self-supervised pretraining. Its training setup, including data splitting, optimizer, and learning rate, is kept consistent with other baselines for fair comparison.

\noindent{\textbf{MLP:}}
As a simple baseline, we implement a multi-layer perceptron (MLP) that directly maps the input neural spike tensor to the target output. Unlike other pretrained models, the MLP is not pretrained in a self-supervised manner and is trained end-to-end directly on the downstream task. For a fair comparison, its training setup, including data splitting and optimization strategy, is kept consistent with that of other methods.

\noindent{\textbf{VAE~\cite{VAE}:}}
To provide a generative modeling baseline, we implement a Variational Autoencoder (VAE) tailored for neural spike data. Given an input spike tensor, we first flatten it into a vector and feed it into a multi-layer perceptron (MLP)-based encoder. The encoder consists of two fully connected layers with Layer Normalization and GELU activations, which output the mean $\mu$ and log-variance $\log \sigma^2$ of the latent distribution. A latent variable $z$ is then sampled via the reparameterization trick, and the decoder, implemented as another MLP, reconstructs the input from the latent representation. The model is optimized with the objective
\begin{equation}
    \mathcal{L} = \mathcal{L}_{\mathrm{recon}} + \beta \cdot \mathcal{L}_{\mathrm{KL}}
\end{equation}
where $\mathcal{L}_{\mathrm{recon}}$ denotes the reconstruction loss and $\mathcal{L}_{\mathrm{KL}}$ denotes the Kullback--Leibler divergence. For a fair comparison, the VAE baseline is trained under the same pretraining protocol as our proposed model, including the same large-scale neural spike dataset, identical training schedule (optimizer, learning rate, and number of epochs), and a fully self-supervised learning paradigm without any label information. We use the AdamW optimizer with cosine annealing learning rate scheduling. The reconstruction objective is implemented using mean squared error (MSE) loss, which is empirically more stable for spike signal reconstruction. A small weighting factor $\beta = 10^{-5}$ is applied to the KL divergence term to balance representation learning and reconstruction fidelity.

\noindent{\textbf{NoMAD~\cite{nomad}:}}
NoMAD trains the dynamical model and decoder on a reference session, and then learns an alignment network in an unsupervised manner on subsequent sessions to map new neural data onto the fixed dynamical model of the reference session. In this work, tailored to downstream tasks, we use 60\% of the data to train the dynamical model and decoder in a supervised manner, followed by 20\% of the data for further unsupervised alignment, and reserve the remaining 20\% for evaluation.

\noindent{\textbf{NDT1~\cite{ye2021ndt}:}}
Neural Data Transformer (NDT) is a non-recurrent framework for modeling neural population activity based on a Transformer encoder architecture. Unlike traditional recurrent neural networks (RNNs), NDT employs self-attention to process neural spike sequences in parallel and captures latent neural dynamics through a masked modeling objective. In this work, we adopt the NDT1 pretrained model provided in~\cite{mtm} for evaluation. The released weights are obtained by retraining NDT1 on their proprietary dataset comprising 34 recording sessions, ensuring a strong and well-optimized baseline. Consistent with our proposed model, we use NDT1 as the encoder and attach a task-specific head composed of a multi-layer perceptron (MLP) for downstream prediction.

\noindent{\textbf{NDT2~\cite{NDT2}:}}
Neural Data Transformer 2 (NDT2) is an extension of the original NDT architecture, designed to address distribution shifts in neural signals across different experiments and subjects through multi-context pretraining. Compared to its predecessor, NDT2 adopts a more powerful Transformer architecture and supports joint training across diverse tasks and recording conditions, enabling the model to learn more generalizable representations of neural population dynamics. In this work, we adopt the NDT2 pretrained model provided in~\cite{mtm} for evaluation. The released weights are obtained by retraining NDT2 on their proprietary dataset consisting of 34 recording sessions. Consistent with our proposed model, we use NDT2 as the encoder and attach a task-specific head composed of a multi-layer perceptron (MLP) for downstream prediction.

\noindent{\textbf{MtM~\cite{mtm}:}}
Multi-task Masking (MtM) is a self-supervised framework for modeling neural population activity, which learns multi-scale representations by alternately performing masking and reconstruction tasks across different spatial scales (single-neuron, intra-region, and inter-region) and temporal dimensions. By introducing learnable prompt tokens for task switching, MtM enables a unified architecture to jointly optimize multiple objectives, including neural activity reconstruction, forward prediction, and inter-region interaction modeling. In this work, we adopt the pretrained MtM model built upon the NDT2~\cite{NDT2} architecture, as provided in~\cite{mtm}. The released weights are obtained by applying their self-supervised training strategy on the IBL repeated-site dataset with 34 recording sessions. Consistent with our proposed model, we use the pretrained model as the encoder and attach a task-specific head composed of a multi-layer perceptron (MLP) for downstream prediction.

\noindent{\textbf{POYO~\cite{poyo}:}}
POYO (Pre-training On manY neurOns) is a unified framework for neural population decoding, which represents individual spike events as discrete tokens consisting of neuron embeddings and timestamps. Built upon the PerceiverIO architecture with rotary positional encoding, POYO enables scalable modeling of neural recordings across animals and sessions. Unlike the self-supervised approaches considered in this work, POYO is trained in a supervised manner. In our experiments, we directly adopt the pretrained POYO-mp model provided by the authors and apply it to downstream tasks via fine-tuning.

\noindent{\textbf{Others:}} 
Other baseline models, including POYO+~\cite{poyo}, POSSM~\cite{POSSM}, NEDS~\cite{neds}, and NDT3~\cite{ndt3}, are not included in our comparison due to the lack of publicly available code or pretrained models. We note that this constraint is common in the literature, and we instead compare against a set of widely adopted and reproducible baselines.

\section{METRICS DESCRIPTION}
\label{app_metrics}
We evaluate the proposed method on both classification and regression tasks using a comprehensive set of metrics. For classification, we adopt: 1) \textbf{Balanced Accuracy}, defined as the average recall across all classes, making it suitable for both binary and multi-class settings, especially under class imbalance, and 2) \textbf{Weighted F1 score}, the harmonic mean of precision and recall, weighted by class support to provide a balanced evaluation across classes. For regression, we report the coefficient of determination ($R^2$), which measures the proportion of variance in the target variable explained by the model. 

\section{ADDITIONAL RESULTS}
\label{add_res}
\subsection{Few-shot Evaluation}
To further evaluate the data efficiency of the proposed model, we conduct few-shot experiments by limiting the amount of labeled data available during training. Specifically, we use only 20\% of the data for training and reserve the remaining 80\% for testing. This setting simulates realistic scenarios in neural data analysis, where obtaining large-scale labeled data is expensive and time-consuming.

The results on M1-CO1 and LICK datasets are summarized in Table~\ref{ACC_fewshot}. Compared to the standard setting, all methods experience performance degradation due to the limited supervision. However, UniBCI consistently outperforms all baselines by a clear margin. On the M1-CO1 dataset, UniBCI achieves a balanced accuracy of 0.782 and a weighted F1 score of 0.781, surpassing the previous best method POYO. On the LICK dataset, UniBCI achieves the highest balanced accuracy of 0.732 and the best weighted F1 score of 0.839. These results demonstrate that UniBCI maintains strong discriminative ability even with severely limited labeled data.

\begin{table}[h]
    \centering
    \caption{Performance comparison on M1-CO1 and LICK datasets.}
    \renewcommand{\arraystretch}{1.3} 
    \resizebox{0.9\textwidth}{!}{
        \begin{tabular}{l|cc|cc}
            \toprule
            \multirow{2}{*}{\textbf{Method}} & \multicolumn{2}{c|}{\textbf{M1-CO1 (Multi-day)}} & \multicolumn{2}{c}{\textbf{LICK}} \\
            \cline{2-5}
                                             & \textbf{Balanced Accuracy} & \textbf{Weighted F1} & \textbf{Balanced Accuracy} & \textbf{Weighted F1} \\
            \midrule
            WF                               & 0.372 $\pm$ 0.048          & 0.415 $\pm$ 0.109          & 0.550                      & 0.505                      \\
            GRU                              & 0.410 $\pm$ 0.056          & 0.435 $\pm$ 0.056          & 0.519                      & 0.509                      \\
            MLP                              & 0.499 $\pm$ 0.029          & 0.496 $\pm$ 0.099          & 0.609                      & 0.811                      \\
            VAE                              & 0.576 $\pm$ 0.069          & 0.572 $\pm$ 0.106          & 0.638                      & 0.820                      \\
            NDT1                             & 0.509 $\pm$ 0.048          & 0.503 $\pm$ 0.110          & 0.660                      & 0.825                      \\
            NDT2                             & 0.725 $\pm$ 0.031          & 0.721 $\pm$ 0.106          & 0.686                      & 0.834                      \\
            MtM                              & 0.727 $\pm$ 0.053          & 0.725 $\pm$ 0.101          & 0.671                      & 0.829                      \\
            POYO                             & \underline{0.762 $\pm$ 0.031} & \underline{0.763 $\pm$ 0.116} & \underline{0.687}          & \underline{0.835}          \\
            \midrule
            UniBCI                           & \textbf{0.782 $\pm$ 0.030} & \textbf{0.781 $\pm$ 0.109} & \textbf{0.732}             & \textbf{0.839}             \\
            \bottomrule
        \end{tabular}
    }
    \label{ACC_fewshot}
\end{table}

\begin{table}[h]
    \centering
    \caption{Performance comparison of different methods.}
    \renewcommand{\arraystretch}{1.2}
    \resizebox{0.90\textwidth}{!}{
        \begin{tabular}{l|ccccc}
            \toprule
            \textbf{Method} & \textbf{Perich T-CO} & \textbf{Perich T-RT} & \textbf{MC-Maze} & \textbf{Area2-Bump} & \textbf{Avg. $R^2$}\\
            \midrule
            WF              & 0.372 $\pm$ 0.048      & 0.415 $\pm$ 0.109      & 0.550            & 0.505               & 0.461          \\
            GRU             & 0.410 $\pm$ 0.056      & 0.435 $\pm$ 0.056      & 0.519            & 0.509               & 0.468          \\
            MLP             & 0.610 $\pm$ 0.029      & 0.509 $\pm$ 0.099      & 0.704            & 0.507               & 0.583          \\
            VAE             & 0.634 $\pm$ 0.069      & 0.511 $\pm$ 0.106      & 0.813            & 0.704               & 0.666          \\
            NDT1            & 0.636 $\pm$ 0.048      & 0.528 $\pm$ 0.110      & 0.862            & \textbf{0.814}      & 0.710          \\
            NDT2            & 0.643 $\pm$ 0.031      & 0.543 $\pm$ 0.106      & \underline{0.867}& 0.793               & \underline{0.712}\\
            MtM             & 0.584 $\pm$ 0.053      & 0.512 $\pm$ 0.101      & \textbf{0.871}   & 0.761               & 0.682          \\
            POYO            & \underline{0.650 $\pm$ 0.031} & \textbf{0.568 $\pm$ 0.116} & 0.846            & 0.742               & 0.702          \\
            \midrule
            UniBCI          & \textbf{0.653 $\pm$ 0.030} & \underline{0.562 $\pm$ 0.109} & 0.846            & \underline{0.798}   & \textbf{0.715} \\
            \bottomrule
        \end{tabular}
    }
    \label{R2_fewshot}
\end{table}

We further evaluate the model on more challenging cross-dataset transfer tasks, as shown in Table~\ref{R2_fewshot}. Under the few-shot setting, the performance gap between methods becomes more pronounced. UniBCI achieves the best performance on Perich T-CO and competitive results on Perich T-RT, while maintaining robust performance on MC-Maze and Area2-Bump. Overall, UniBCI achieves the highest average score of 0.715, outperforming all competing methods. Notably, while some baselines exhibit performance drops under limited data, UniBCI shows more stable degradation, indicating stronger robustness.


\subsection{Ablation Study}

We conduct an ablation study on two key design choices of the proposed model: positional encoding and token grouping strategy. Specifically, we compare STPE (spatio-temporal absolute positional encoding) with RoPE (rotary positional encoding), and evaluate two grouping strategies: Uniform Group and Direct Group. For grouping, Uniform Group first evenly partitions channels (or units) into a predefined number of areas (\textit{Area Number}), and pads each group to a fixed size (\textit{Area Size}) when necessary, ensuring balanced and consistent token structures. In contrast, Direct Group directly pads all channels to the maximum length and then splits them into groups according to \textit{Area Number}, which may result in some groups containing mostly or entirely zero-padding tokens. As shown in Table~\ref{tab:ablation_design}, STPE consistently provides more stable performance across tasks than RoPE. Although RoPE achieves comparable classification results on M1-CO1, it leads to clear performance degradation on the regression tasks MC-Maze and Area2-Bump, indicating that explicit temporal indexing is more suitable for neural spike representations. For grouping strategy, Uniform Group generally outperforms Direct Group by maintaining more regular token distributions and reducing redundant padded regions. Direct Group introduces irregular structures and may generate empty groups, which weakens representation consistency and harms overall robustness, despite showing competitive results on Area2-Bump. Overall, the combination of STPE and Uniform Group achieves the best trade-off between accuracy, robustness, and efficiency, validating its effectiveness for neural signal modeling.

\begin{table}[h]
\centering
\caption{Ablation study on positional encoding and grouping strategy.}
\label{tab:ablation_design}
\renewcommand{\arraystretch}{1.2}
\resizebox{\linewidth}{!}{
\begin{tabular}{l|cc|cc}
\toprule
\multirow{2}{*}{\textbf{Variants}} 
& \multicolumn{2}{c|}{\textbf{M1-CO1 (Multi-day)}} 
& \multirow{2}{*}{\textbf{MC-Maze ($R^2$)}} 
& \multirow{2}{*}{\textbf{Area2-Bump ($R^2$)}} \\
\cline{2-3}
& \textbf{Balanced Accuracy} 
& \textbf{Weighted F1} 
& & \\
\midrule
STPE + Uniform Group & \textbf{0.895} & \textbf{0.897} & \textbf{0.890} & 0.890 \\
RoPE + Uniform Group & 0.893 & 0.894 & 0.850 & 0.840 \\
STPE + Direct Group  & 0.834 & 0.833 & 0.879 & \textbf{0.892} \\
\bottomrule
\end{tabular}
}
\end{table}

\subsection{Sensitivity Analysis}
To further evaluate the robustness of the proposed model, we conduct a comprehensive sensitivity analysis on key tokenization hyperparameters, including Interval Size ($t$), Area Size ($C_a$), and Area Number ($A$), with results summarized in Table \ref{tab:appendix_sensitivity}. We observe that the model is highly sensitive to temporal resolution: an excessively small interval ($t=2$) leads to a significant performance drop (e.g., Balanced Accuracy 0.670 on M1-CO1), likely due to fragmentation of continuous neural dynamics that impairs global dependency modeling within limited attention contexts, while an overly large interval ($t=20$) also degrades performance by discarding critical high-frequency temporal information; in contrast, the default setting ($t=10$) achieves the best trade-off between temporal fidelity and efficiency. For Area Size ($C_a$), increasing $C_a$ from 8 to 32 consistently improves performance, suggesting enhanced modeling of population-level features within each token, whereas further increasing to $C_a=64$ yields marginal gains or slight degradation, indicating that overly large areas may introduce redundancy and obscure distinct neural contributions, highlighting the importance of mesoscopic-scale representations. Furthermore, the interaction between Area Number ($A$) and sliding window size ($w$) exhibits dataset-dependent behavior: for example, a compact configuration ($A=2, w=4$) achieves the best performance on Area2-Bump ($R^2=0.901$), while more complex tasks such as MC-Maze benefit from larger $A$, suggesting that different neural dynamics favor different spatial granularities. Overall, these results demonstrate that while the model is sensitive to temporal granularity ($t$), it remains robust across a wide range of area-based configurations ($A$, $C_a$), validating the effectiveness and adaptability of the proposed Area-wise Sliding Window Attention in modeling diverse neural data structures.
\begin{table}[h]
    \centering
    \caption{Comprehensive sensitivity analysis of tokenization hyperparameters.}
    \label{tab:appendix_sensitivity}
    \renewcommand{\arraystretch}{1.2}
    \resizebox{\textwidth}{!}{
        \begin{tabular}{cccc|cc|c|c}
            \toprule
            \multirow{2}{*}{$t$} & \multirow{2}{*}{$w$} & \multirow{2}{*}{$A$} & \multirow{2}{*}{$C_a$} & \multicolumn{2}{c|}{\textbf{M1-CO1 (Multi-day)}} & \multirow{2}{*}{\textbf{MC-Maze ($R^2$)} } & \multirow{2}{*}{\textbf{Area2-Bump ($R^2$)} } \\
            \cline{5-6}
             &  &  &  & \textbf{Balanced Accuracy} & \textbf{Weighted F1} & & \\
            \midrule
            2 & 10 & 8 & 32 & 0.670 & 0.672 & 0.790 & 0.812 \\
            5 & 10 & 8 & 32 & 0.853 & 0.856 & 0.846 & 0.866 \\
            10 & 10 & 8 & 32 & \textbf{0.895} & \textbf{0.897} & \textbf{0.890} & \underline{0.890} \\
            20 & 10 & 8 & 32 & 0.856 & 0.859 & 0.867 & 0.850 \\
            \midrule
            10 & 10 & 8 & 8 & 0.824 & 0.826 & 0.836 & 0.875 \\
            10 & 10 & 8 & 16 & 0.878 & 0.880 & 0.874 & 0.871 \\
            10 & 10 & 8 & 64 & 0.872 & 0.871 & 0.860 & 0.882 \\
            \midrule
            10 & 4 & 2 & 32 & 0.837 & 0.835 & 0.863 & \textbf{0.901} \\
            10 & 6 & 4 & 32 & 0.861 & 0.862 & 0.885 & 0.880 \\
            10 & 18 & 16 & 32 & 0.848 & 0.853 & 0.874 & 0.882 \\
            \bottomrule
        \end{tabular}
    }
\end{table}

\subsection{The Effectiveness of Pretraining}
As shown in Table \ref{tab:ablation_scratch}, we compared the performance of UniBCI trained from scratch versus our proposed model with pretraining. It is observed that UniBCI (From-scratch) performs significantly worse across all generalization tasks. This performance gap stems from the inclusion of textual features in our architecture. Without a large-scale pretraining phase, the model fails to effectively align the high-dimensional textual semantic features with the neural signal characteristics. In the absence of this alignment, the textual features may act as meaningless noise or biased inputs during the supervised training stage, leading to suboptimal feature representation and degraded decoding accuracy. Thus, the pretraining stage is crucial for grounding the language priors within the neural decoding process.

\begin{table}[hbp]
    \centering
    \caption{The effect of pretraining on model performance across different tasks. $\pm$ represents SD over sessions.}
    \renewcommand{\arraystretch}{1.3} 
    \resizebox{1.0\textwidth}{!}{
        \begin{tabular}{l|cc|cc|cc}
            \toprule
            \textbf{Generalization} & \multicolumn{2}{c|}{\textbf{New Session}} & \multicolumn{2}{c|}{\textbf{New Subject}} & \multicolumn{2}{c}{\textbf{New Species}} \\
            \midrule
            \multirow{2}{*}{\textbf{Method}} & \multicolumn{2}{c|}{\textbf{M1-CO1 (Multi-day)}} & \multicolumn{2}{c|}{\textbf{LICK}} & \multicolumn{2}{c}{\textbf{PPC-FINGER}} \\
            \cline{2-7}
            & \textbf{Balanced Accuracy} & \textbf{Weighted F1} & \textbf{Balanced Accuracy} & \textbf{Weighted F1} & \textbf{Balanced Accuracy} & \textbf{Weighted F1} \\
            \midrule
            UniBCI (From-scratch) & 0.799 & 0.804 & 0.689 $\pm$ 0.034 & 0.842 $\pm$ 0.024 & 0.835 $\pm$ 0.041 & 0.829 $\pm$ 0.042 \\
            UniBCI  & \textbf{0.895} & \textbf{0.897} & \textbf{0.744 $\pm$ 0.019} & \textbf{0.859 $\pm$ 0.007} & \textbf{0.967 $\pm$ 0.020} & \textbf{0.960 $\pm$ 0.023} \\
            \bottomrule
        \end{tabular}
    }
    \label{tab:ablation_scratch}
\end{table}

\section{VISUALIZATION}
\subsection{Context Representations}
As shown in Figure~\ref{fig:text_embed}, we investigate the semantic consistency and discriminability of text representations to better understand the role of natural language as a modality alignment bridge in cross-dataset and cross-species neural pretraining. A key challenge in building unified feature spaces for heterogeneous neural signals, such as spike trains recorded from different brain regions, tasks, and species, lies in achieving robust alignment across highly non-stationary experimental conditions. To address this, we introduce structured natural language descriptions as a shared semantic interface and analyze their embedding properties using a pretrained text encoder such as MiniLM-L6-v2~\cite{minilmv2}. Specifically, we construct structured prompts based on dataset metadata from both pretraining datasets (M1-CO1, M1-CO2, Pac-Man, HPC-HG, LICK) and downstream tasks, following the template:
\begin{figure}[h]
    \centering
    \includegraphics[width=0.95\linewidth]{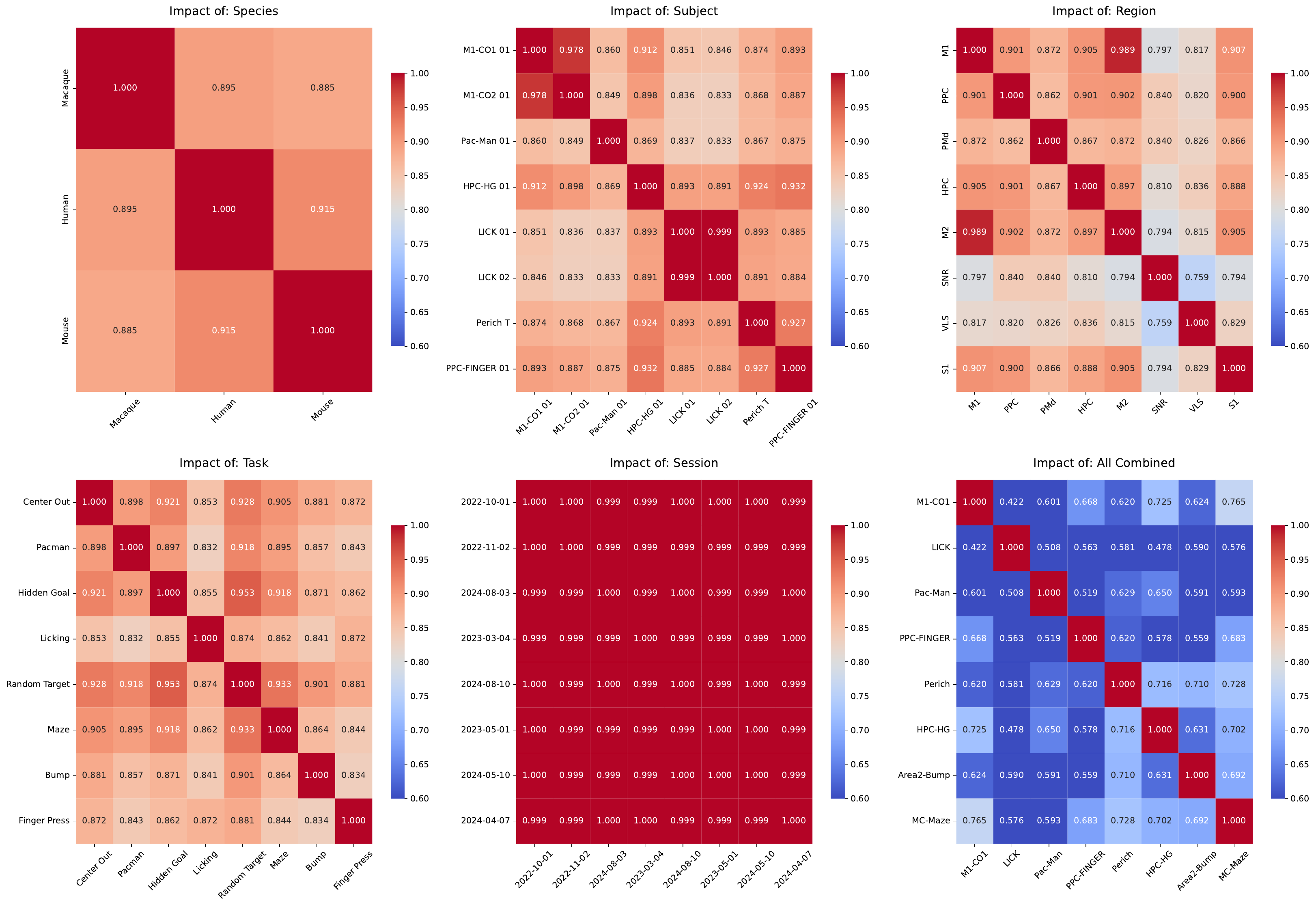}
    \caption{Semantic consistency and discriminability of text representations. Heatmaps show cosine similarities under controlled variable settings, where one attribute (species, subject, region, task, or session) is varied while others are fixed, revealing strong invariance to subject and session and clear discriminative structure for task and region.}
    \label{fig:text_embed}
\end{figure}

\textit{"Invasive spike signals of {\color{red}[Species]} species ({\color{blue}[Dataset] [Subject]}) in the {\color{green}[Region]} brain region during the {\color{purple}[Task]} task under session {\color{orange}[Session/Date]}"}

We then apply a controlled variable strategy, where we fix all but one factor and systematically vary a single attribute, including species, subject, brain region, task, and session date. Cosine similarities between corresponding text embeddings are computed to generate heatmaps that visualize semantic variation across different dimensions. The results show strong semantic stability with respect to session and subject variations, where similarity values remain consistently above 0.99, indicating that the text encoder effectively smooths inter-day and inter-subject variability and captures invariant experimental structure. In contrast, task and brain region dimensions exhibit clear discriminative gradients, with similarities ranging between 0.85 and 0.95, suggesting that the text encoder can effectively distinguish functional differences such as M1 versus hippocampus or center-out versus Pacman tasks, thereby providing informative task-level guidance. Furthermore, when multiple factors are jointly varied, the similarity decreases, in some cases below 0.6, demonstrating strong global differentiation capability. Overall, these results indicate that structured textual descriptions preserve high semantic consistency while still maintaining sufficient discriminability to distinguish heterogeneous neural datasets, thereby validating natural language as an effective and generalizable bridge for cross-dataset neural representation learning and knowledge transfer.

\subsection{Spike Token Representations}
\begin{figure}[h]
    \centering
    \includegraphics[width=0.95\linewidth]{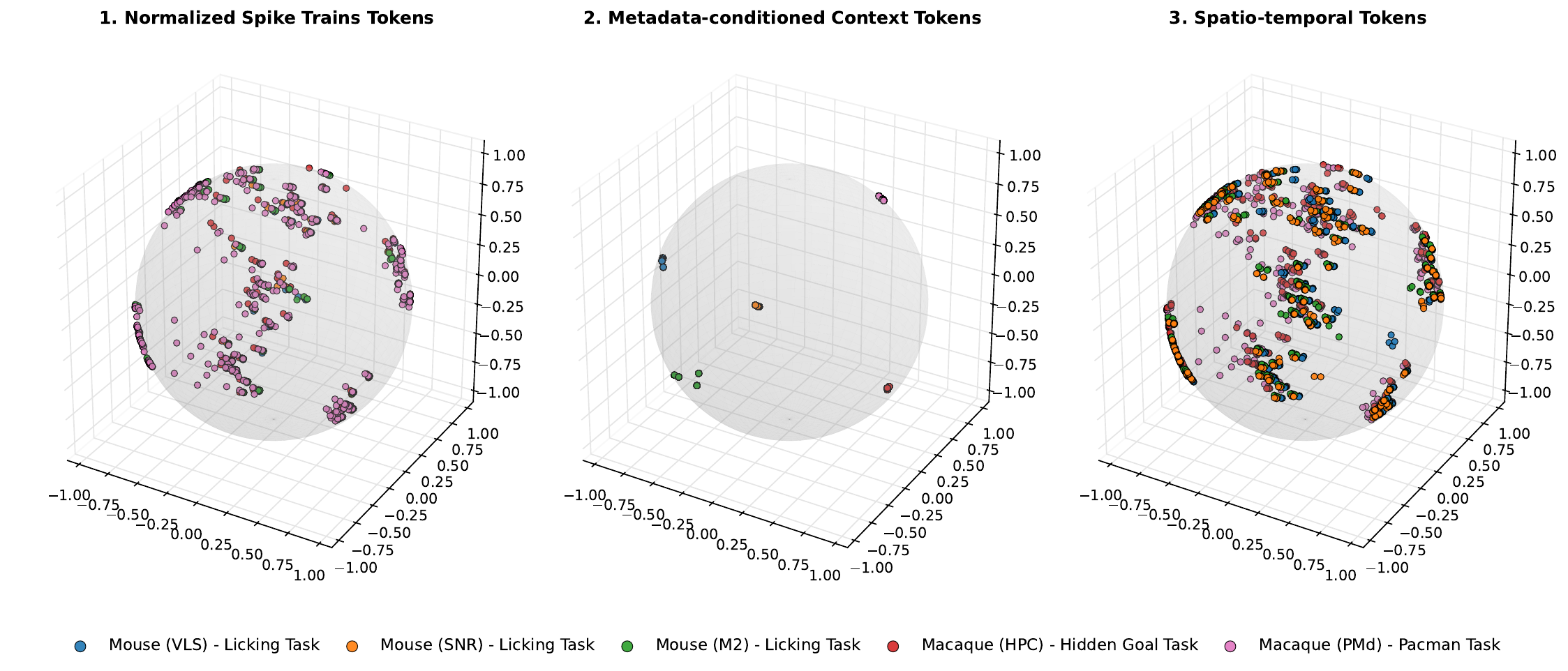}
    \caption{Token distributions on a hyperspherical space across stages, showing how metadata progressively structures neural representations from entangled spike tokens to organized spatiotemporal embeddings.}
    \label{fig:metadata}
\end{figure}
As illustrated in Fig.~\ref{fig:metadata}, we visualize token representations at different stages using hyperspherical embedding projections to examine how metadata injection reshapes the feature space for cross-species and cross-task neural signals. When tokens are derived solely from raw spike trains, their embeddings are highly entangled and uniformly dispersed on the hypersphere, with substantial overlap across species such as mouse and macaque as well as different brain regions, indicating limited ability to capture domain-specific structure and weak robustness under domain shift. By introducing structured metadata, including species, subject identity, brain region, and task, as context embeddings, the model constructs a more organized representation space where tokens are distributed around distinct contextual anchors, forming separable clusters that reflect underlying experimental conditions and provide a semantic prior for neural signal modeling. After integrating contextual information, the resulting spatiotemporal tokens exhibit a more structured distribution characterized by both global coverage and local compactness, with improved uniformity on the hypersphere while preserving meaningful intra-domain variation, suggesting that metadata acts as a conditioning signal that regularizes representation learning and enhances cross-domain generalization. Overall, metadata injection transforms neural representations from entangled, low-discriminative embeddings into a structured semantic space on the hypersphere, enabling a shift from purely data-driven fitting toward context-aware representation alignment.

\subsection{Encoder Representation}
\begin{figure}[h]
    \centering
    \includegraphics[width=0.9\linewidth]{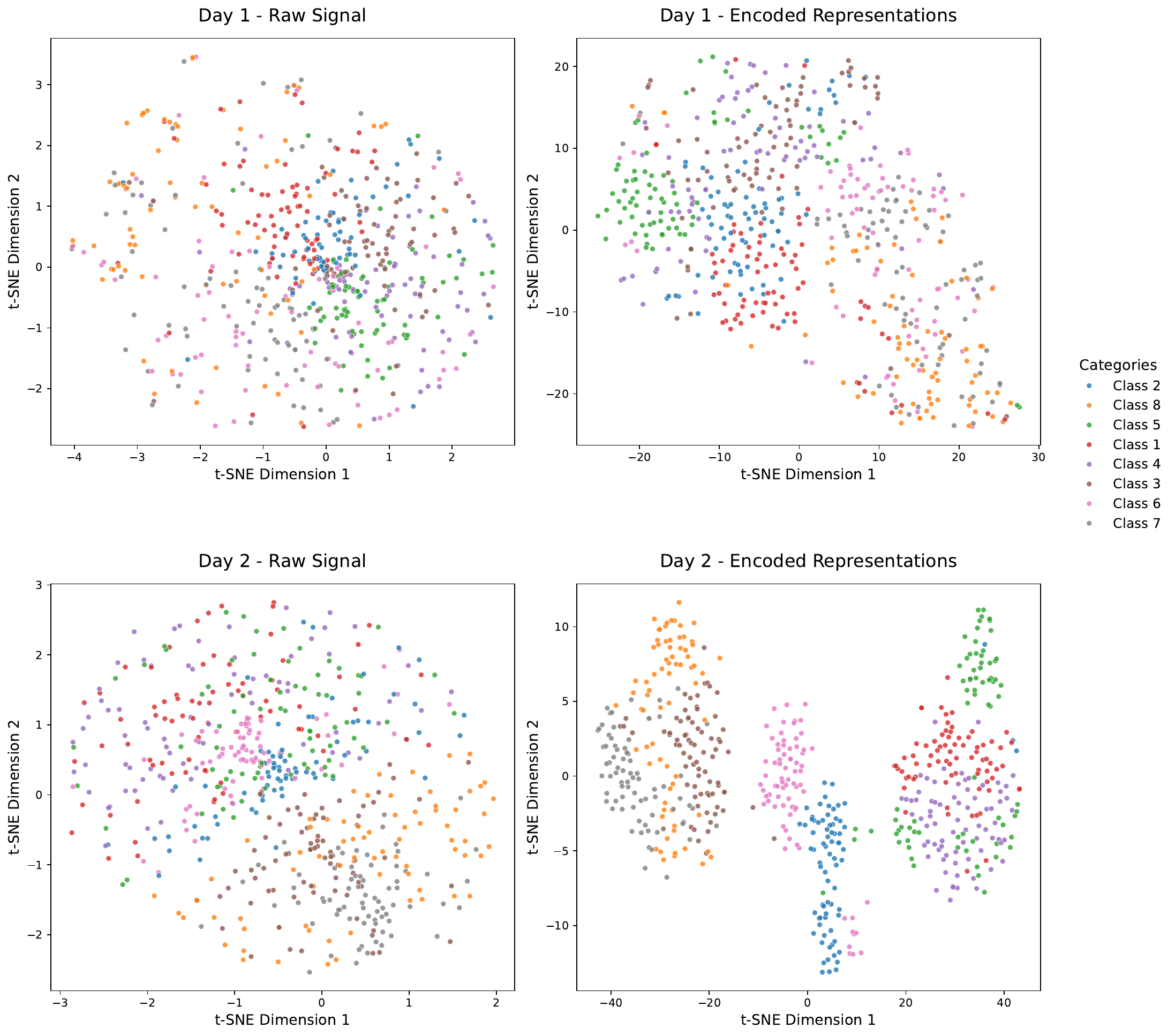}
    \caption{t-SNE visualization of feature representations on M1-CO1, comparing raw neural signals (left) and encoded representations (right).}
    \label{fig:visual}
\end{figure}
To qualitatively evaluate the discriminative capability of the learned neural representations, we conduct a t-SNE visualization on the M1-CO1 macaque dataset, which consists of 232-channel spike recordings from the M1 cortex. We select data from two unseen days comprising six independent sessions that are not included during pretraining, and consider an eight-class center-out movement classification task. As illustrated in Figure~\ref{fig:visual}, we compare the distribution of raw neural signals with the learned representations produced by our model. In the raw signal space, samples from different movement directions are highly entangled across both Day 1 and Day 2, exhibiting no clear class boundaries and overlap among categories. In contrast, in the learned representation space, the encoded features form more compact and well-separated clusters, with clear distinctions between different classes. This result demonstrates that, despite not being exposed to these sessions during pretraining, the proposed model is able to extract discriminative patterns from high-dimensional, sparse, and noisy neural spike sequences that are strongly correlated with movement direction. These findings highlight the effectiveness of the proposed approach in learning robust and transferable neural representations, as well as its strong generalization capability across unseen sessions.

\section{Limitations}
\label{limitations}
Despite the strong empirical performance and generalization ability demonstrated across diverse datasets, several limitations of the current framework should be acknowledged. First, the model relies on pre-defined spike train normalization (e.g., temporal binning and fixed channel grouping) and a structured tokenization pipeline, which impose spatial assumptions on data representation. While this design enables efficient unification of heterogeneous datasets, it may restrict flexibility when handling more complex or irregular recordings, such as variable-density electrode layouts or continuous-time neural events. Second, although UniBCI incorporates contextual metadata embeddings, it remains fundamentally limited to spike-based modalities and does not explicitly support multimodal neural data integration. This constrains its ability to leverage complementary information from other sources, such as neuroimaging (e.g., fMRI, calcium imaging) or behavioral streams, which are increasingly important for comprehensive brain modeling. Future work will therefore focus on developing adaptive, data-driven tokenization schemes and extending the framework toward unified multimodal representation learning.

\newpage

\end{document}